\documentclass{article}

\usepackage[preprint]{neurips}
\usepackage{amsmath}
\usepackage{amssymb}
\usepackage{booktabs}
\usepackage{microtype}
\usepackage{graphicx}
\usepackage{subcaption}
\usepackage{placeins}
\usepackage[hidelinks]{hyperref}
\usepackage{url}

\makeatletter
\renewcommand{\@noticestring}{}
\makeatother
\title{HYPE-EDIT-1: An Effective-Cost and Reliability Benchmark for Image Editing}

\author{Wing Chan \\ Sourceful Ltd \\ \texttt{wing@sourceful.com} \\
\And Richard Allen \\ Sourceful Ltd \\ \texttt{rich@sourceful.com}}

\begin{document}
\maketitle

\begin{abstract}
Public demos of image editing models are typically best-case samples; real workflows pay for retries and review time. We introduce HYPE-EDIT-1, a 100-task benchmark of reference-based marketing/design edits with binary pass/fail judging. For each task we generate 10 independent outputs to estimate per-attempt pass rate, pass@10, expected attempts under a retry cap, and an effective cost per successful edit that combines model price with human review time. We release 50 public tasks and maintain a 50-task held-out private split for server-side evaluation, plus a standardized JSON schema and tooling for VLM and human-based judging. Across the evaluated models, per-attempt pass rates span 34--83\% and effective cost per success spans \$0.66--\$1.42. Models that have low per-image pricing are more expensive when you consider the total effective cost of retries and human reviews.
\end{abstract}

\section{Introduction}
Generative image editing models are increasingly used in product marketing,
brand design, and creative production. In these settings, reliability matters as
much as peak quality: a model that occasionally produces a perfect edit but
frequently fails imposes significant time and cost overhead. Existing evaluations
often emphasize best-case outputs, obscuring how many attempts are required to
obtain a usable result. HYPE-EDIT-1 targets this gap by quantifying both
reliability and effective cost on real-world editing tasks that require precise
changes to reference images.
Effective cost is important because current models are marketed around their
cost per image, with models that are cheaper per image seen as 'more affordable'.
This benchmark shows that once differences in the number of retries are
accounted for, effective cost is not proportional to the per-image cost.
For example, in our results a model priced at \$0.03/image still reaches
about \$1.42 per successful edit once retries and review are accounted for, while
a higher-priced model can be cheaper per success.

\begin{figure*}[t]
  \centering
  \begin{subfigure}{0.95\textwidth}
    \centering
    \includegraphics[width=\linewidth]{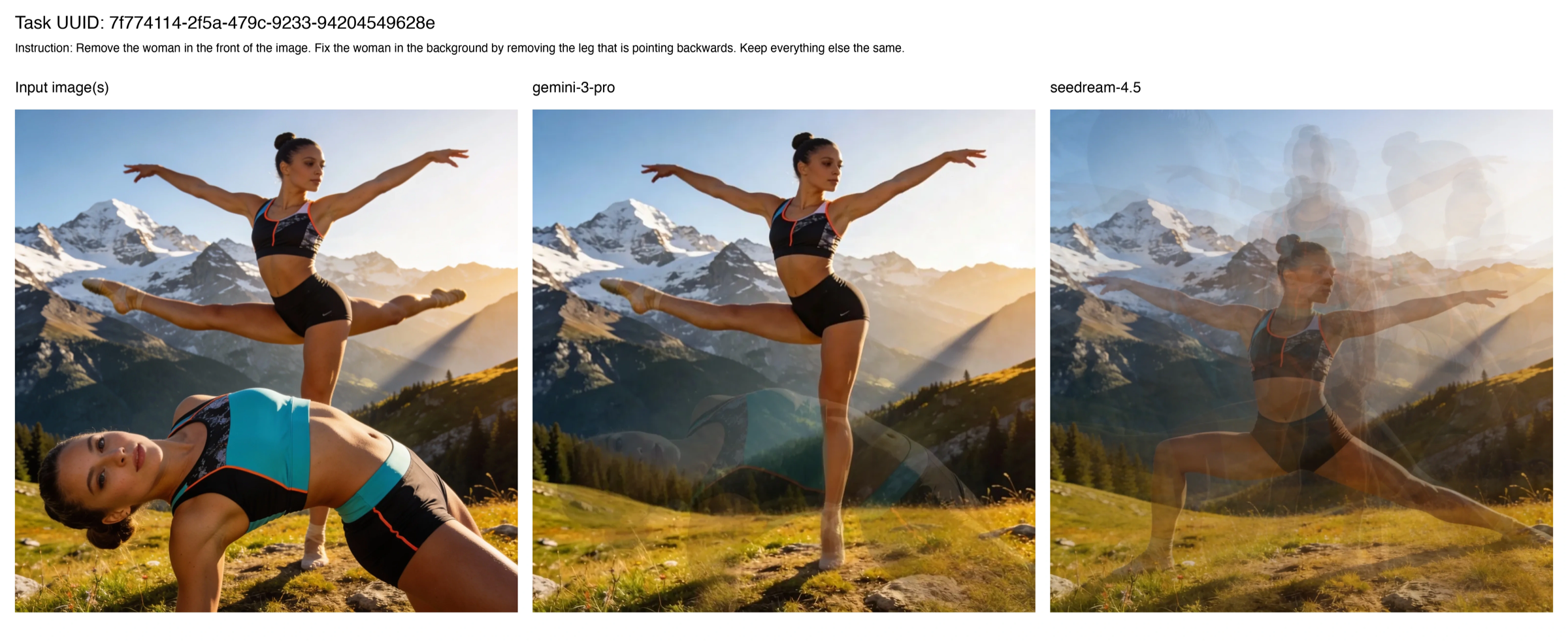}
    \caption{Task 7f774114 (public).}
  \end{subfigure}

  \begin{subfigure}{0.95\textwidth}
    \centering
    \includegraphics[width=\linewidth]{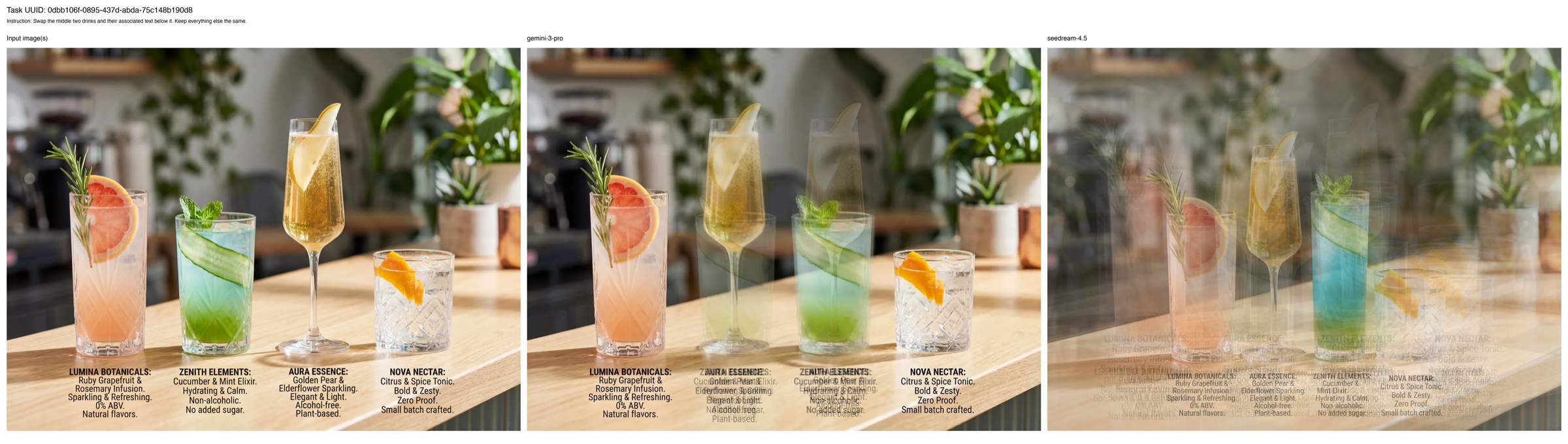}
    \caption{Task 0dbb106f (public).}
  \end{subfigure}
  \caption{Qualitative comparisons between Gemini 3 Pro Preview (Nano Banana Pro) and Seedream 4.5. Each
  panel shows the input reference(s) alongside 10-sample collages for each
  model. The collage view makes it easy to see how repeated attempts at the same
  task can diverge, even for narrowly scoped edits such as remove or swap where
  minimal variation is expected. In practice, these collages often show partial
  instruction compliance or structure drift that would be hidden by selecting a
  single best output.}
  \label{fig:qualitative-examples}
\end{figure*}

Figure~\ref{fig:qualitative-examples} shows qualitative comparisons between
Gemini 3 Pro Preview (Nano Banana Pro) and Seedream 4.5 on two public tasks. Each comparison image
includes the input reference(s) on the left and a collage of 10 independent
outputs per model (via the comparison image generator), highlighting differences
in instruction adherence, consistency, and variation across retries.

HYPE-EDIT-1 is built around the notion that practical workflows demand
consistency. For each task, the same prompt is executed repeatedly and scored
independently, allowing us to estimate a success probability. These repeated
trials underpin metrics that reflect the actual user experience of retrying a
model until it succeeds.

\begin{figure*}[t]
  \centering
  \captionsetup[subfigure]{font=footnotesize,justification=raggedright,singlelinecheck=false}
  \begin{subfigure}{0.32\textwidth}
    \centering
    \includegraphics[width=\linewidth,height=0.16\textheight,keepaspectratio]{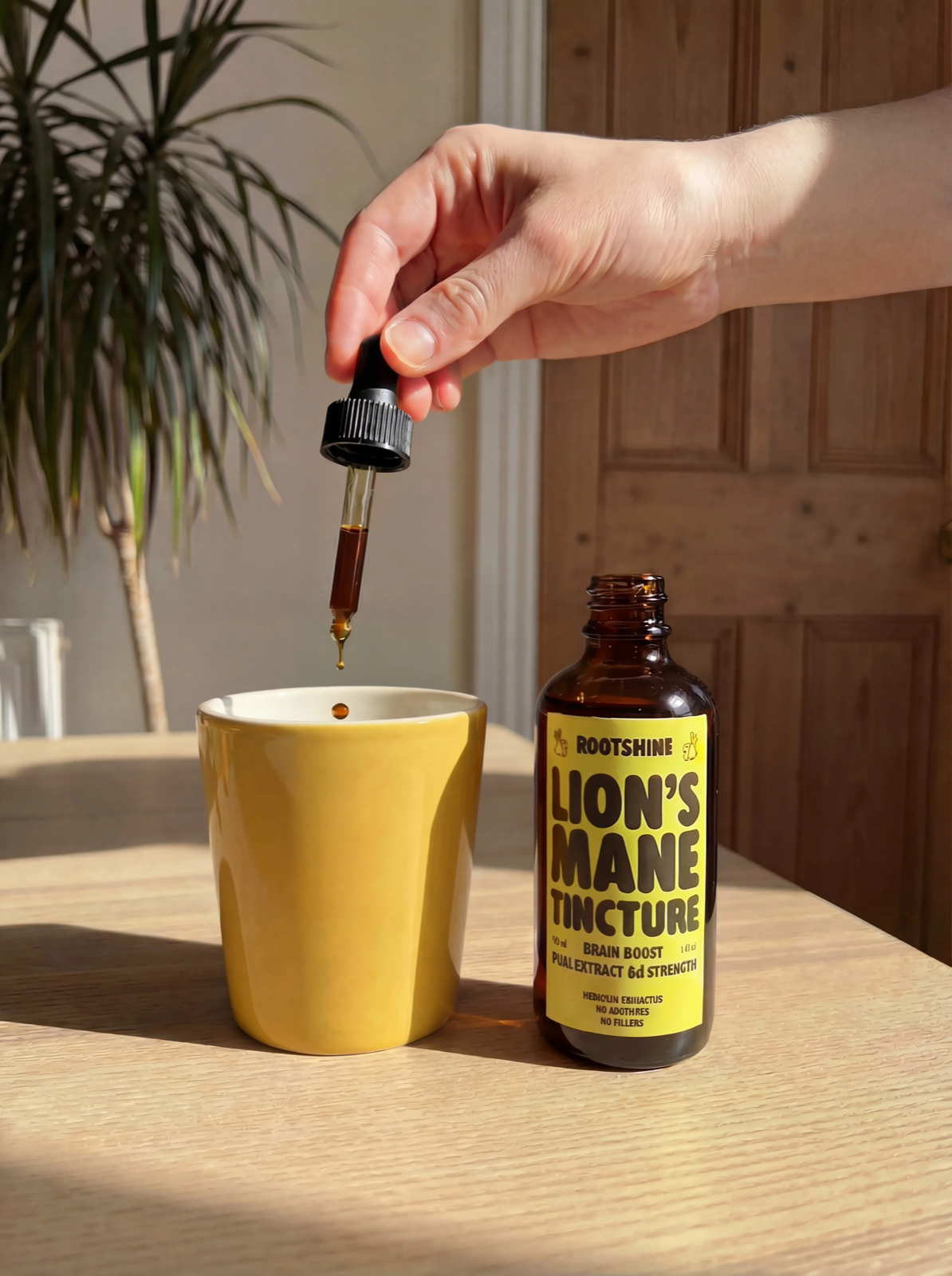}
    \caption{\textbf{change}: Add a handle to the mug. Keep everything else the same.}
  \end{subfigure}
  \hfill
  \begin{subfigure}{0.32\textwidth}
    \centering
    \includegraphics[width=\linewidth,height=0.16\textheight,keepaspectratio]{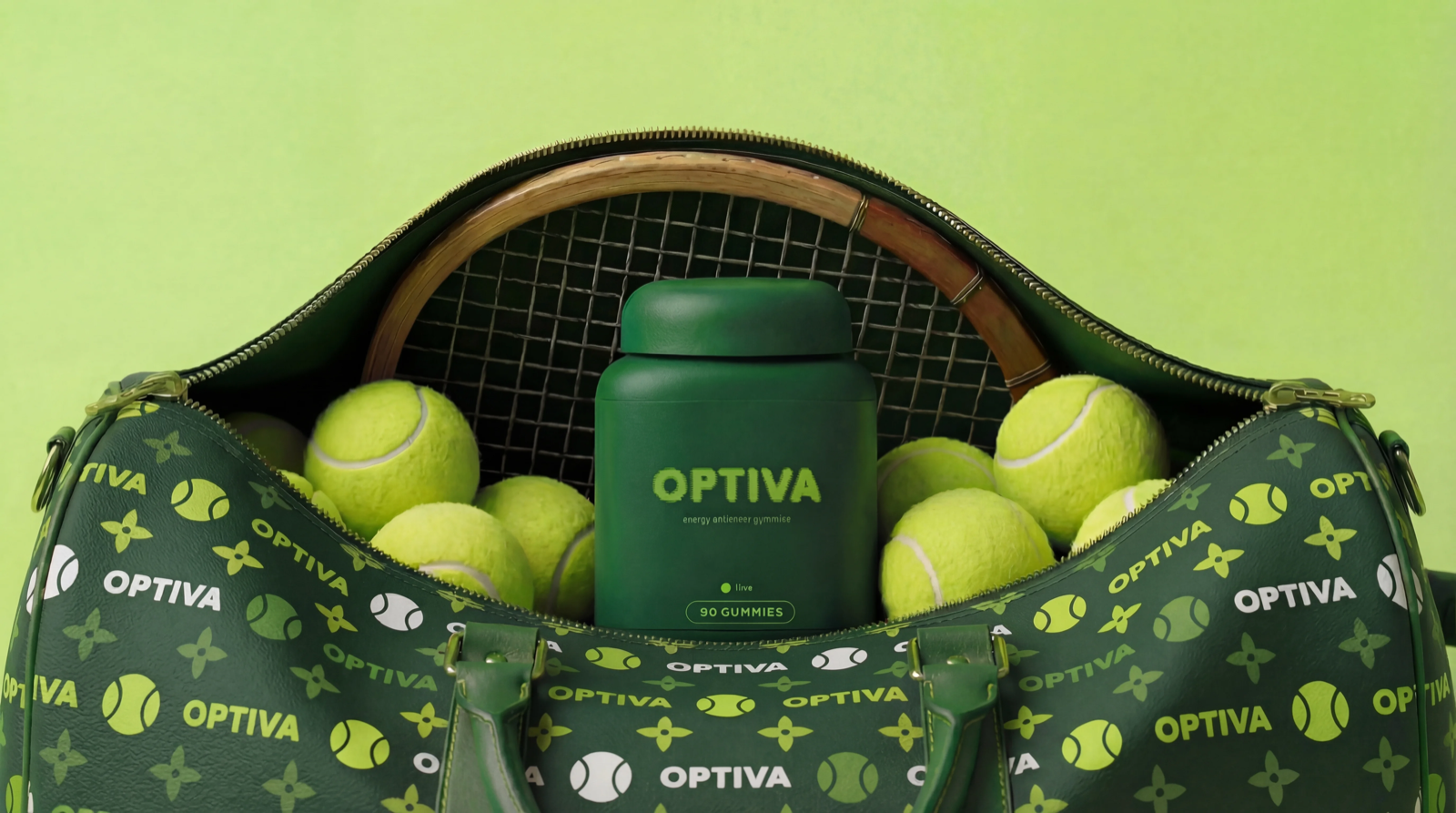}
    \caption{\textbf{remove}: Remove all the artwork elements from the bag apart from its base color. Keep everything else the same.}
  \end{subfigure}
  \hfill
  \begin{subfigure}{0.32\textwidth}
    \centering
    \includegraphics[width=\linewidth,height=0.16\textheight,keepaspectratio]{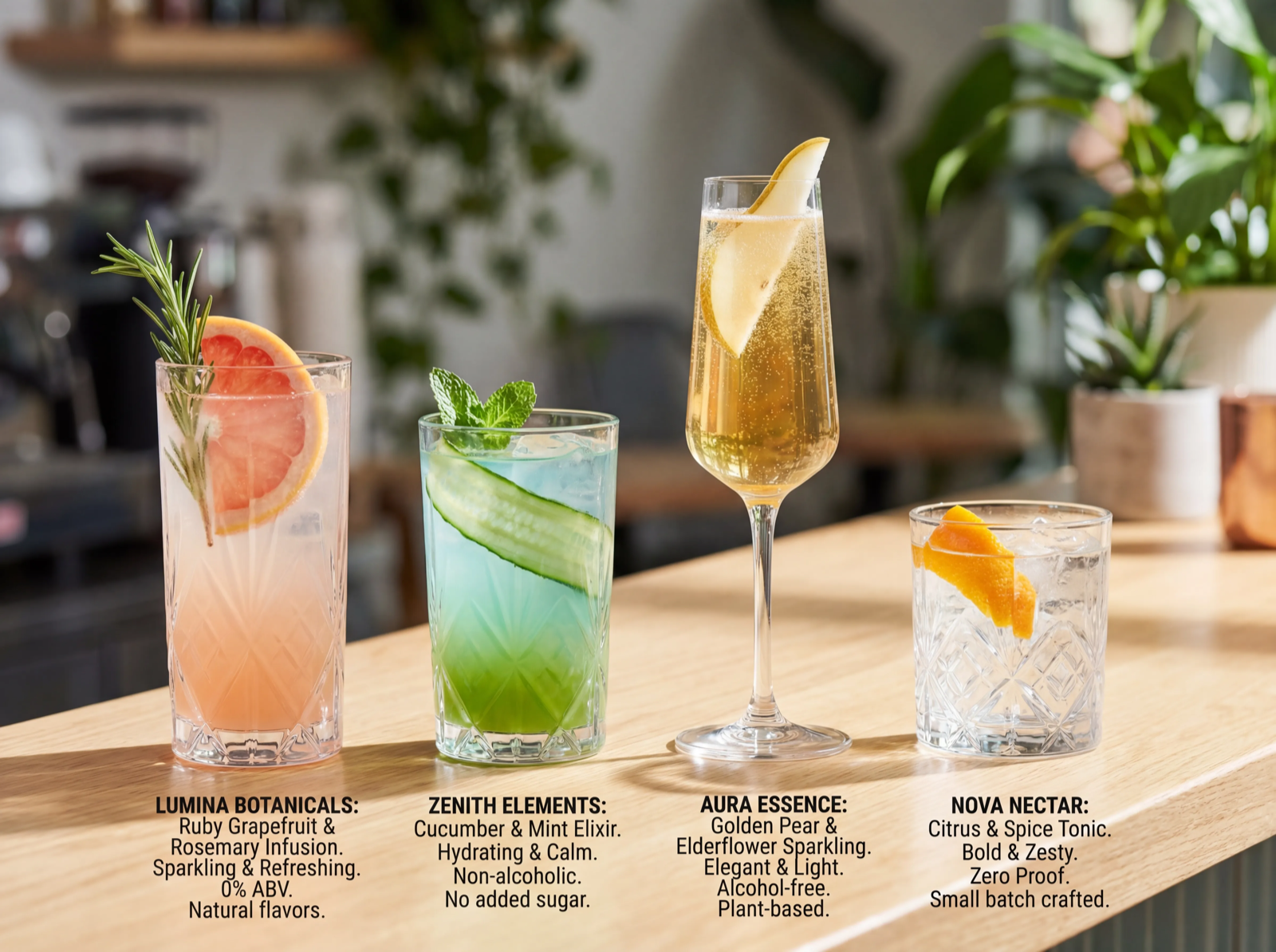}
    \caption{\textbf{restructure}: Swap the middle two drinks and their associated text below it. Keep everything else the same.}
  \end{subfigure}

  \begin{subfigure}[t]{0.48\textwidth}
    \centering
    \begin{subfigure}{0.48\linewidth}
      \centering
      \includegraphics[width=\linewidth,height=0.16\textheight,keepaspectratio]{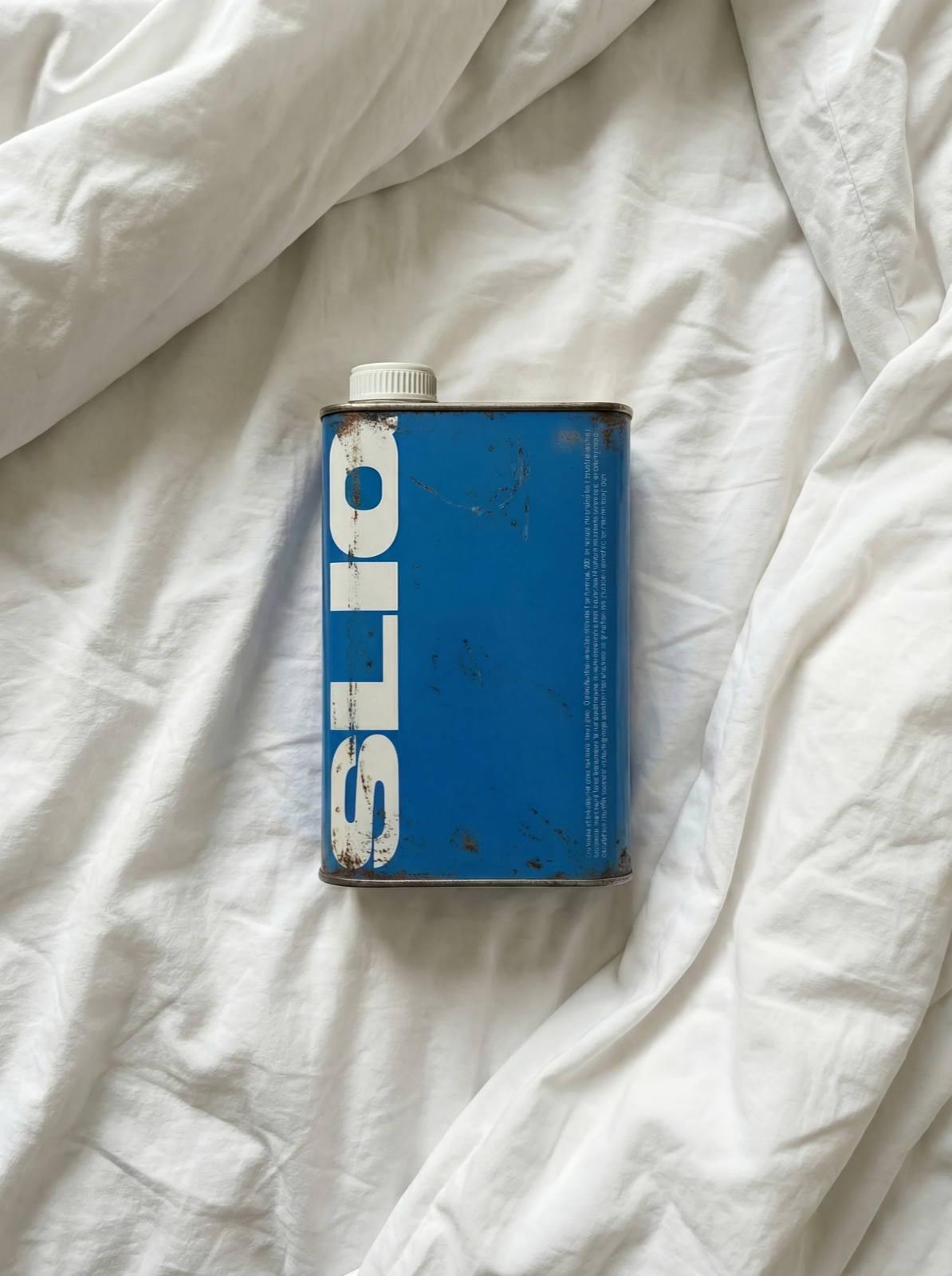}
    \end{subfigure}
    \hfill
    \begin{subfigure}{0.48\linewidth}
      \centering
      \includegraphics[width=\linewidth,height=0.16\textheight,keepaspectratio]{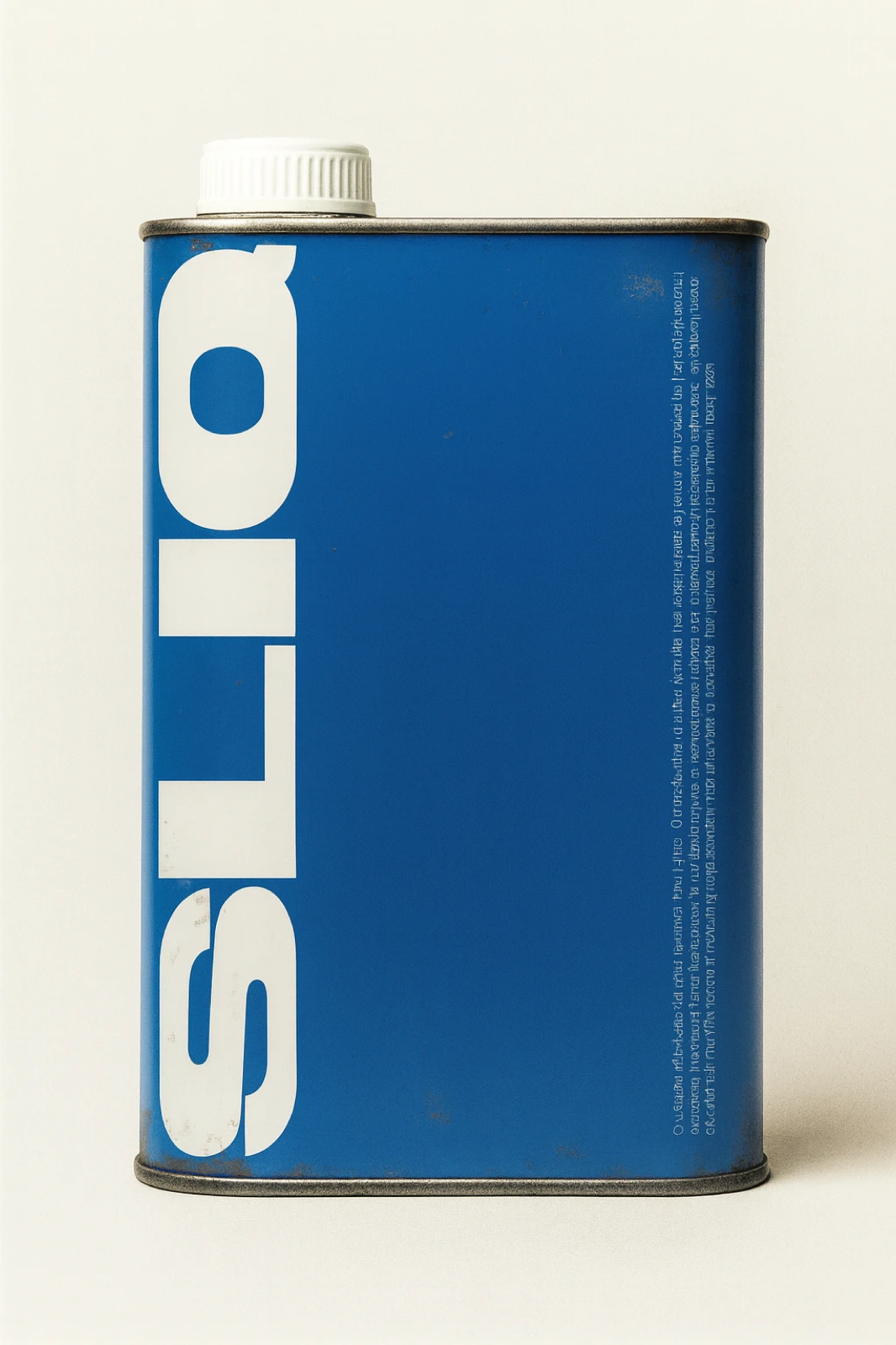}
    \end{subfigure}
    \caption{\textbf{enhance}: Fix the product artwork in the first image [Image 1] by using the second image [Image 2] as the artwork reference. Change the artwork but keep everything else the same.}
  \end{subfigure}%
  \hfill%
  \begin{subfigure}[t]{0.48\textwidth}
    \centering
    \includegraphics[width=\linewidth,height=0.16\textheight,keepaspectratio]{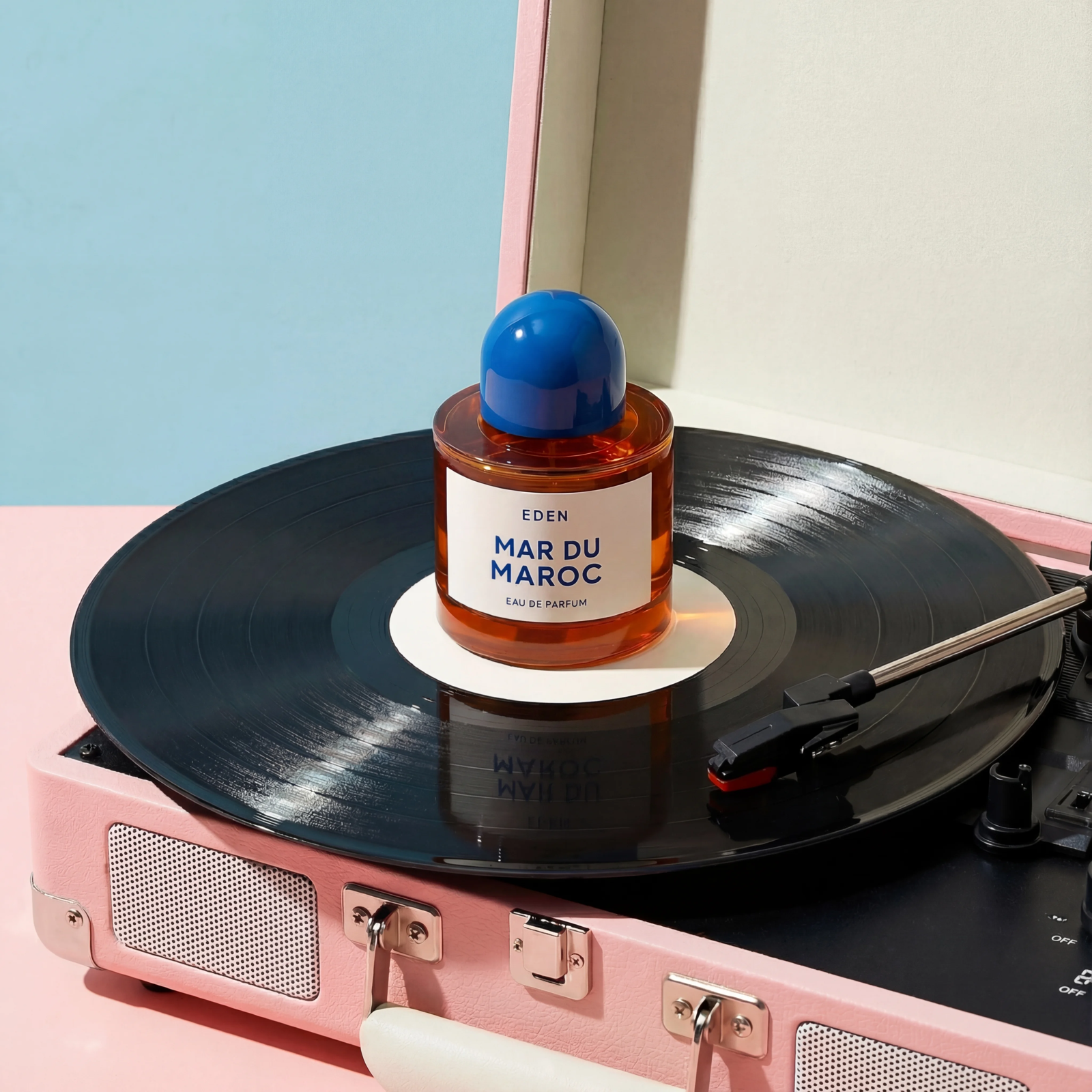}
    \caption{\textbf{change}: Change the color of the vinyl inner circle from white to blue to exactly match the blue top of the perfume bottle. Keep everything else the same.}
  \end{subfigure}
  \caption{Example tasks from HYPE-EDIT-1 showing diverse editing instructions
  across different task types: change, remove, restructure, and enhance. Each
  panel shows the input reference image(s) with the instruction and task category.}
  \label{fig:task-examples}
\end{figure*}

Figure~\ref{fig:task-examples} shows five example tasks from the benchmark,
illustrating the diversity of editing instructions and task types.

\FloatBarrier

\section{Related Work}
\subsection{Instruction-following image editing models}
Early image editing systems relied on domain-specific supervision, masks, or
per-image optimization, which limited usability for open-ended edits. Conditional
GAN editors such as Pix2Pix and CycleGAN enabled paired or unpaired translation
but lacked open-ended instruction control \citep{isola2017pix2pix,zhu2017cyclegan}.
Diffusion and inversion-based methods such as SDEdit \citep{meng2022sdedit},
Null-Text Inversion \citep{mokady2023nulltext}, and Prompt-to-Prompt
\citep{hertz2022prompt} made it possible to preserve input structure while
changing semantics from text prompts, but often required masks or carefully
tuned latent initializations. Blended Diffusion introduced localized edits with
masking and CLIP guidance \citep{avrahami2022blended}, while SINE explored
single-image editing without explicit masks \citep{sine2023}. InstructPix2Pix
shifted the paradigm by training on synthetic instruction-image pairs,
demonstrating that a single model can generalize to diverse edits without
per-example optimization \citep{brooks2023instructpix2pix}. Imagic further
showed that fine-tuning on a single input image enables non-rigid edits but at
the cost of per-image optimization time \citep{imagic2023}. Subsequent work
introduced instruction-tuned models and curated datasets such as MagicBrush,
which adds human-edited examples and multi-turn sessions \citep{magicbrush2023},
and MGIE, which uses multimodal language models to refine editing instructions
\citep{mgie2024}. These advances motivate benchmarks that test general
instruction following while accounting for practical constraints like iteration
cost and reliability.

\subsection{Datasets and benchmarks}
Benchmarks for text-guided image editing have evolved from paired datasets with
ground-truth targets to reference-free evaluation. GIER and MagicBrush provide
paired edits and enable direct pixel or feature comparisons, but they cannot
capture the one-to-many nature of creative edits \citep{gier2020,magicbrush2023}.
TEdBench introduced a small, challenging test set for open-ended edits with
human preference evaluation \citep{imagic2023}. EditVal, HATIE, and I2EBench
scale evaluation through VLM-based metrics that are calibrated against human
judgments \citep{editval2023,hatie2025,i2ebench2024}, while SpotEdit extends
evaluation to multi-modal instructions with additional reference images and
hallucination tests \citep{spotedit2025}. Exemplar-guided datasets such as
PaintByExample and DreamEdit highlight reference-image conditioning for editing
\citep{paintbyexample2023,dreamedit2023}. Table~\ref{tab:benchmark-comparison}
summarizes key differences between representative benchmarks and HYPE-EDIT-1.

\begin{table*}[t]
  \caption{Comparison of HYPE-EDIT-1 with representative image editing
  benchmarks. Scales and attributes are reported from published descriptions
  where available.}
  \label{tab:benchmark-comparison}
  \centering
  \small
  \setlength{\tabcolsep}{4pt}
  \resizebox{\textwidth}{!}{%
  \begin{tabular}{l c c c c l}
    \toprule
    Benchmark & Scale & Edit scope & Evaluation & Repeats & Split \\
    \midrule
    GIER & 6,179 images & local/global & GT outputs & No & Train/val/test \\
    MagicBrush & 5,313 sessions & multi-turn & GT outputs & No & Train/val/test \\
    TEdBench & 100 pairs & non-rigid & Human prefs & No & Test-only \\
    EditVal & 648 tasks & 13 edit types & VLM metrics & No & Public + held-out \\
    HATIE & 18,226 images & broad & Composite auto & No & Reported splits \\
    I2EBench & 2,000 images & 16 dims & Learned judge & No & Public \\
    SpotEdit & 500 samples & text+ref & GT + checks & No & Public \\
    HYPE-EDIT-1 & 100 tasks & marketing edits & Human panel (majority) + VLM check & Yes (10x) & Public + private \\
    \bottomrule
  \end{tabular}%
  }
\end{table*}

\subsection{Model-based evaluation}
With open-ended edits, learned metrics based on VLMs have become the
standard alternative to pixel-level comparison. CLIP-based similarity scores,
attribute classifiers, and composite metrics are commonly used to judge whether
an edit satisfies the instruction while preserving unrelated content. EditVal,
HATIE, and I2EBench each report correlations between automatic scores and human
preferences, supporting the use of VLM judges at scale
\citep{editval2023,hatie2025,i2ebench2024}. HYPE-EDIT-1 follows this trend by
providing a VLM judge example (Gemini 3 Flash (gemini-3-flash-preview)) and a human-judge web UI to
support manual review.

\subsection{Reliability and stochastic evaluation}
Generative models are inherently stochastic, but most image editing benchmarks
evaluate a single output per task. In other domains, pass@k-style metrics
capture the probability of success over multiple attempts, reflecting how users
iterate in practice \citep{chen2021codex}. Image generation systems similarly
rely on sampling and re-ranking, yet reliability metrics remain underreported.
HYPE-EDIT-1 addresses this gap by evaluating each task over 10 independent
attempts and reporting pass-at-10, expected attempts, and effective cost.

\subsection{Workflow considerations and positioning}
Creative workflows value iteration speed, predictability, and alignment with
domain constraints such as marketing requirements or brand consistency. While
prior benchmarks emphasize either scale or fine-grained edit categories,
HYPE-EDIT-1 focuses on real-world marketing and design edits and explicitly
measures reliability and effective cost. The benchmark is smaller than recent
large-scale suites, but it provides repeated-trial evaluation, a public/private
split to mitigate leakage, and both automated and human judging workflows,
positioning it as a complementary, reliability-focused benchmark
\citep{hartmann2024marketing,liu2023design}.

\section{Benchmark Design}
\subsection{Task construction}
HYPE-EDIT-1 consists of curated, real-world image editing tasks drawn from
marketing and design workflows. Each task includes one or two reference images
and a concise instruction describing the desired edit. The tasks are not
puzzle-like; instead, they require concrete modifications such as removing an
object, adjusting the orientation of a product, or restructuring a layout.

\subsection{Task types and scale}
The benchmark contains 100 tasks split evenly between public and private sets.
Task types include \texttt{change}, \texttt{remove}, \texttt{restructure}, and
\texttt{enhance}. Most tasks use a single input image, while a smaller subset
requires combining or transferring content from two images. Input image
resolutions range from 2048 to 5504 pixels on the long edge.

\begin{table}[t]
\caption{HYPE-EDIT-1 dataset summary.}
\label{tab:dataset}
\centering
\begin{tabular}{lr}
\toprule
Statistic & Value \\
\midrule
Total tasks & 100 \\
Public tasks & 50 \\
Private tasks & 50 \\
Single-image tasks & 89 \\
Multi-image tasks & 11 \\
Task types (change/remove/restructure/enhance) & 50/21/17/12 \\
Resolution range (width x height) & 2048--5504 x 1728--5504 \\
Repeats per task & 10 \\
\bottomrule
\end{tabular}
\end{table}

\subsection{Data schema and access}
Tasks are stored as JSON entries with fields: \texttt{task\_id},
\texttt{instruction}, \texttt{task\_type}, \texttt{input\_images}, and target
\texttt{width} and \texttt{height}. Reference images are hosted at a stable CDN
path keyed by the task identifier. The public split is released in the
repository at \url{https://www.github.com/sourceful-official/hype-edit-1-benchmark},
while the private split is held for server-side evaluation to mitigate training
contamination.

\section{Evaluation Protocol}
\subsection{Repeated trials}
For each model and task, HYPE-EDIT-1 generates $K=10$ independent candidate
outputs. This produces 1,000 outputs per model for the full benchmark. The
repeated-trial setup allows the benchmark to estimate a per-task success
probability rather than relying on a single example.

\subsection{Pass or fail judging}
Each output is judged against the task instruction by a panel of five human
raters who vote PASS or FAIL without seeing which model produced the image; the
majority vote determines the final label. We also run a visual-language model
(VLM) judge for alignment using a deterministic scoring prompt and a fixed
threshold for PASS/FAIL. The VLM judge (Gemini 3 Flash (gemini-3-flash-preview))
is used as a check and agrees with the human majority in roughly 80\% of cases.
The VLM is more critical and often fails on subtle changes that remain
acceptable to human reviewers. The repository includes a Gemini 3 Flash
(gemini-3-flash-preview) judge example and a human-judge web UI.

\subsection{Reliability and effective cost}
Let $y_{t,k} \in \{0, 1\}$ denote whether the $k$-th attempt on task $t$ is
successful. With $T$ tasks and $K$ repeats, overall reliability is

\begin{equation}
R = \frac{1}{T K} \sum_{t=1}^{T} \sum_{k=1}^{K} y_{t,k}.
\end{equation}

We report Pass Rate (P@1) as the probability of success on the first attempt and
Pass@10 (P@10) as the probability of at least one success across the 10 attempts.

We report expected attempts and effective cost using the same retry-cap model
implemented in our internal analysis script; the equations here allow
reproduction.

\subsection{Cost model}
The benchmark cost model combines (1) the model's per-candidate generation cost,
(2) an estimated human review cost per candidate, and (3) the expected number of
attempts required for a task. For model $m$, the per-attempt cost is

\begin{equation}
C_{attempt} = C_{model}(m) + C_{review}.
\end{equation}

We estimate review cost with a default hourly rate of \$50 and 20 seconds of
inspection time per image,

\begin{equation}
C_{review} = (50 / 3600) \cdot 20 \approx 0.278.
\end{equation}

For each task $t$, we estimate a pass rate $p_t$ from the full $K=10$ candidates
and compute the expected number of attempts with a maximum retry budget $A=4$
(reflecting a practical user retry limit). The expected attempts and success
probability under this cap are

\begin{equation}
E_t =
\begin{cases}
\frac{1 - (1 - p_t)^A}{p_t} & \text{if } p_t > 0, \\
A & \text{if } p_t = 0,
\end{cases}
\quad
S_t = 1 - (1 - p_t)^A.
\end{equation}

We then aggregate across tasks to compute pass@4 and effective cost per success
using the retry-cap model:

\begin{equation}
p@4 = \frac{1}{T} \sum_t S_t,
\quad
E = \frac{1}{T} \sum_t E_t.
\end{equation}

\begin{equation}
C_{eff} = \frac{E \cdot C_{attempt}}{p@4}.
\end{equation}

Table~\ref{tab:model-costs} lists the per-candidate model costs used in our
analysis, which serve as the inputs to $C_{model}(m)$.

\begin{table}[t]
  \caption{Per-candidate model costs used in our analysis (USD).}
  \label{tab:model-costs}
  \centering
  \begin{tabular}{lr}
    \toprule
    Model & Cost per candidate (\$) \\
    \midrule
    gemini-3-pro-preview & 0.134 \\
    seedream-4.5 & 0.04 \\
    seedream-4.0 & 0.03 \\
    gpt-image-1.5 & 0.17 \\
    flux-2-max & 0.10 \\
    riverflow-2-b1 & 0.15 \\
    qwen-image-edit-2511 & 0.03 \\
    \bottomrule
  \end{tabular}
\end{table}

\subsection{Image models}
We evaluate a mix of current flagship and preview image models used for
instruction-following edits. Model identifiers match the benchmark
configuration.
\begin{itemize}
  \item \textbf{gemini-3-pro-preview}: Gemini 3 Pro Preview (Nano Banana Pro),
  released November 2025 by Google; 2K resolution outputs.
  \item \textbf{seedream-4.5}: ByteDance Seedream 4.5 flagship image model; 2K
  resolution outputs.
  \item \textbf{seedream-4.0}: ByteDance's previous flagship image model.
  \item \textbf{gpt-image-1.5}: OpenAI's flagship image model; 1K-1.5K resolution and
  limited aspect ratios.
  \item \textbf{flux-2-max}: Black Forest Labs flagship image model; 2K outputs
  and additional reasoning support.
  \item \textbf{riverflow-2-b1}: Sourceful Riverflow 2 (Beta) pre-release model
  targeted for Q1 2026.
  \item \textbf{qwen-image-edit-2511}: Alibaba's latest flagship Qwen Image Edit
  model, released December 2025; 2K image resolution.
\end{itemize}

\subsection{Hype Gap}
We report the \emph{Hype Gap} (Best-of-10 Uplift) as the difference, in
percentage points, between pass@10 and pass@1. Let $p@1$ be the success rate on
the first attempt and $p@10$ be the probability of achieving at least one
success within 10 attempts. Then

\begin{equation}
\text{HypeGap} = p@10 - p@1.
\end{equation}

Lower Hype Gap indicates consistent performance between a model's typical
outputs and its best-of-10 results, while a larger Hype Gap implies that strong
edits are possible but less reliable in practical use.

\section{Benchmark Usage}
The code repository, including task files and tooling, is available at
\url{https://www.github.com/sourceful-official/hype-edit-1-benchmark}. The
benchmark release includes the public task set, a Gemini 3 Flash
(gemini-3-flash-preview) judge example implementation, and a human-judge web UI.
The code is licensed under the MIT license; the tasks and reference imagery are
licensed under CC BY 4.0 and were created by Sourceful, which grants the rights
to use these assets under that license. We define a standard directory layout
keyed by dataset, model name, and task identifier to support consistent
reporting and longitudinal tracking of reliability improvements over time.

\section{Results and Discussion}
HYPE-EDIT-1 is designed as a living benchmark. The latest model evaluations are
reported alongside the benchmark release and are updated as new models appear.
Because the benchmark emphasizes repeated trials and cost-adjusted metrics, it
is well suited to analyze trade-offs between high-cost, high-reliability models
and lower-cost models that require more retries.

\subsection{Combined Dataset Results}
\begin{figure*}[t]
  \centering
  \begin{subfigure}{0.49\textwidth}
    \centering
    \includegraphics[width=\linewidth,height=0.22\textheight,keepaspectratio]{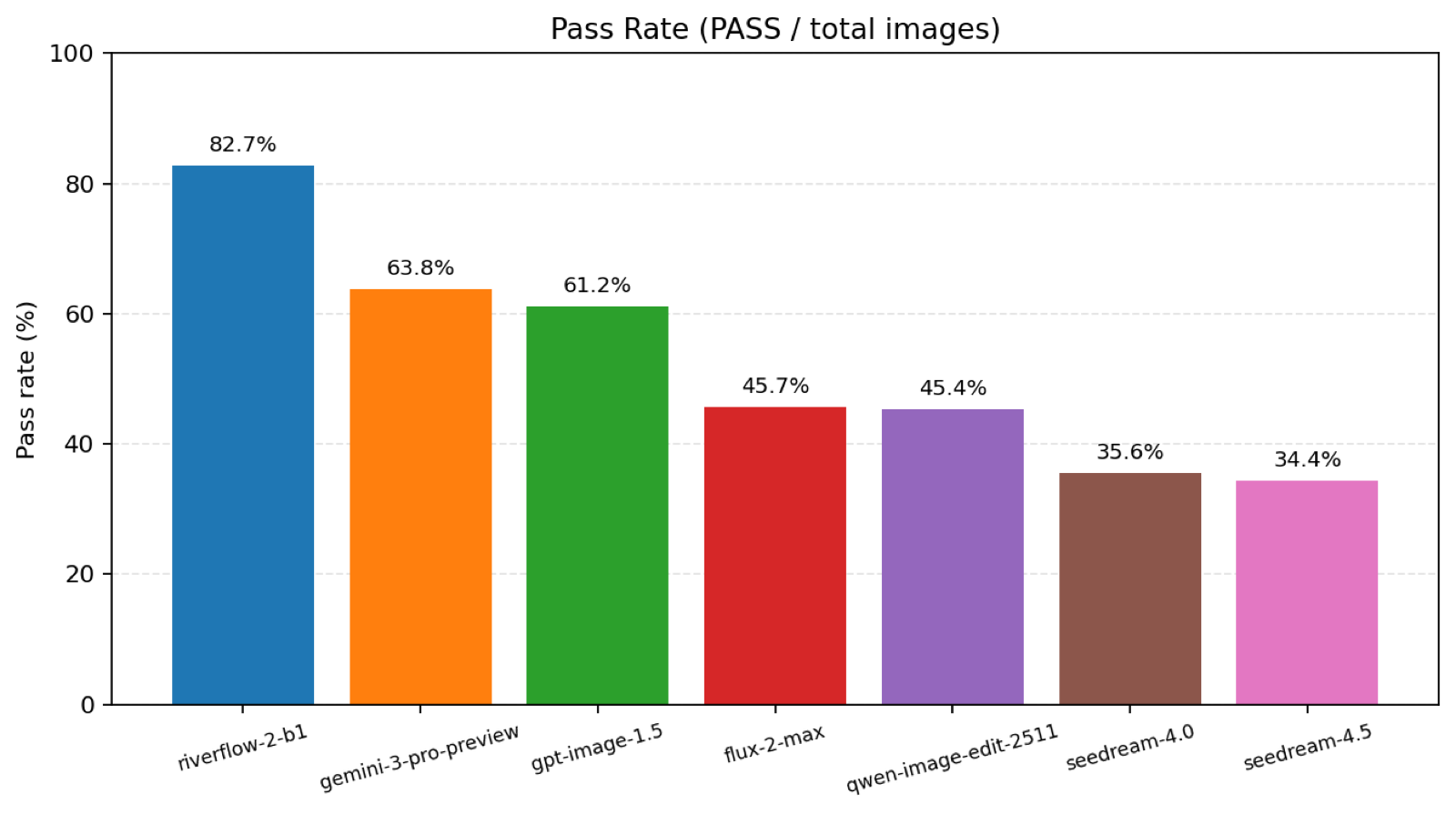}
    \caption{Pass rate (combined).}
  \end{subfigure}
  \hfill
  \begin{subfigure}{0.49\textwidth}
    \centering
    \includegraphics[width=\linewidth,height=0.22\textheight,keepaspectratio]{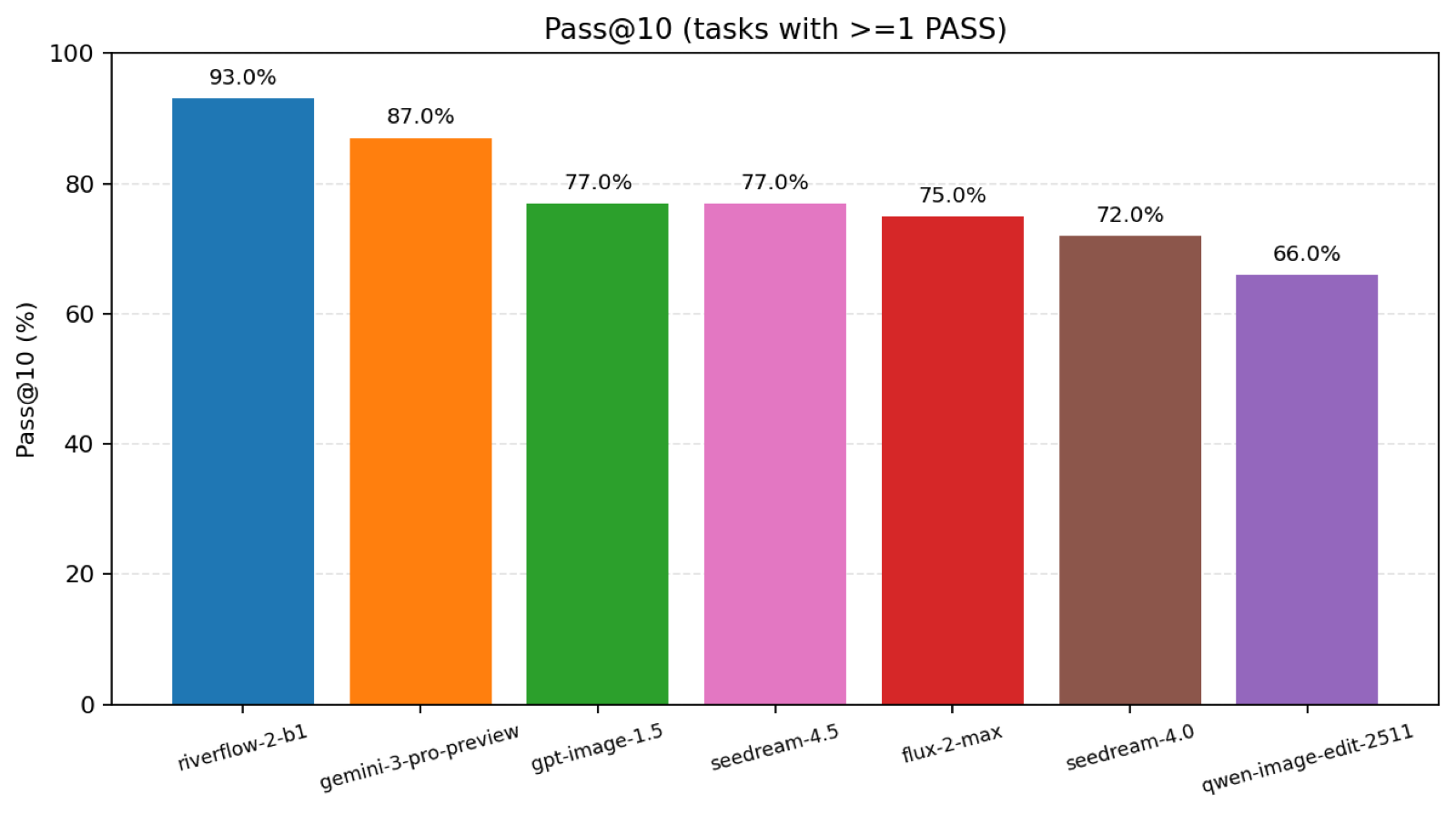}
    \caption{Pass at 10 (combined).}
  \end{subfigure}

  \begin{subfigure}{0.49\textwidth}
    \centering
    \includegraphics[width=\linewidth,height=0.22\textheight,keepaspectratio]{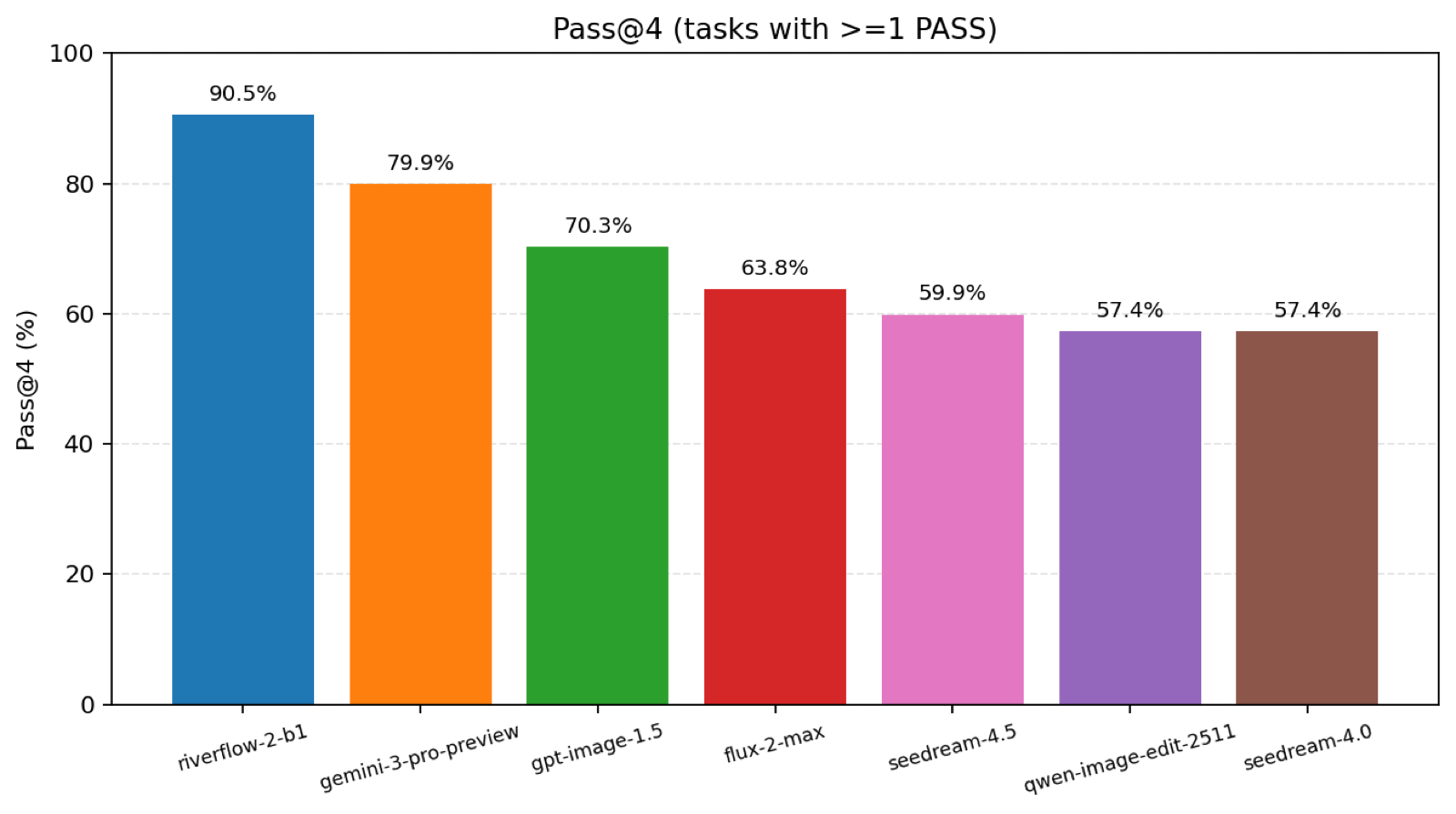}
    \caption{Pass at 4 (combined).}
  \end{subfigure}
  \hfill
  \begin{subfigure}{0.49\textwidth}
    \centering
    \includegraphics[width=\linewidth,height=0.22\textheight,keepaspectratio]{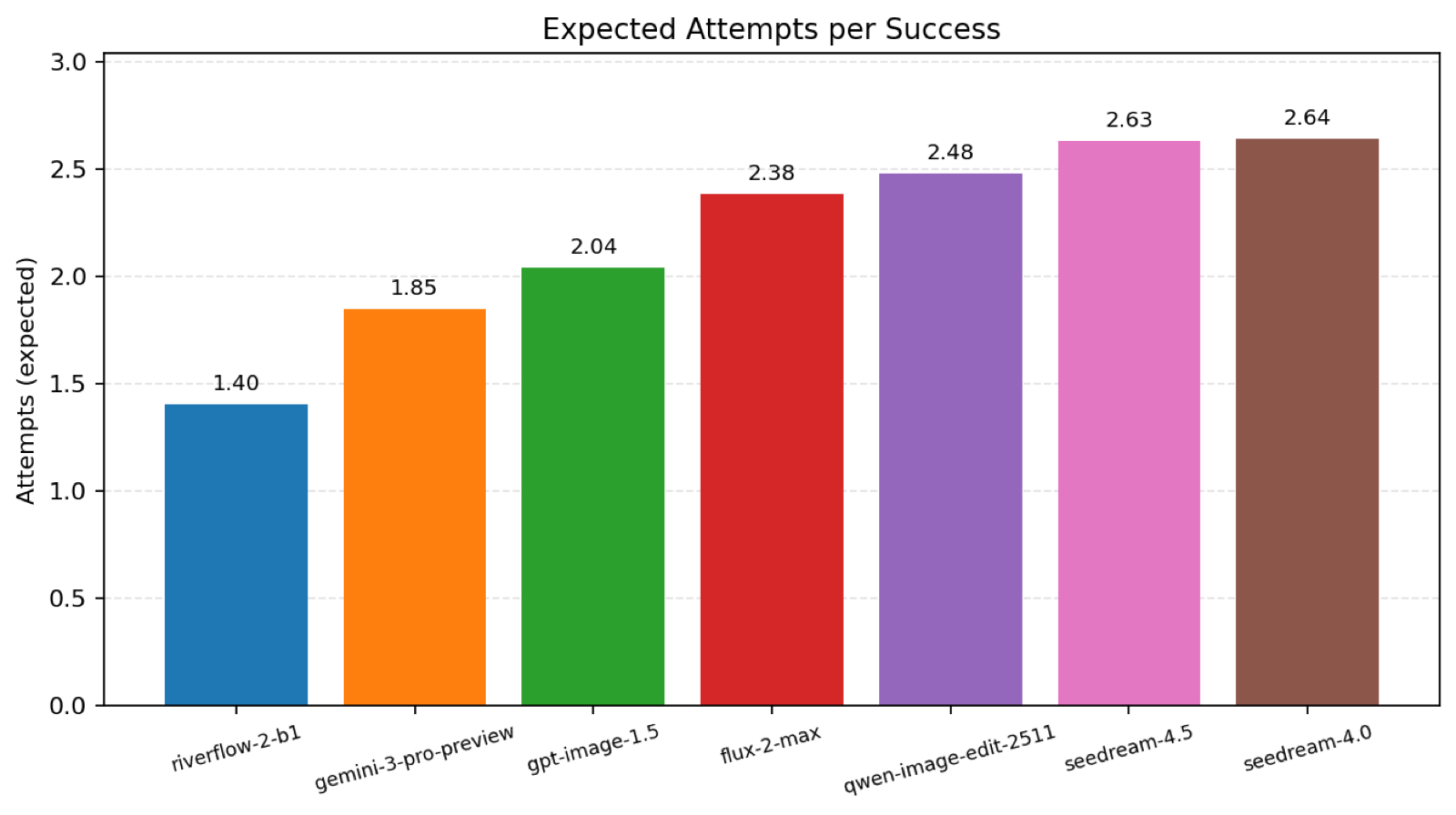}
    \caption{Expected attempts (combined).}
  \end{subfigure}

  \begin{subfigure}{0.49\textwidth}
    \centering
    \includegraphics[width=\linewidth,height=0.22\textheight,keepaspectratio]{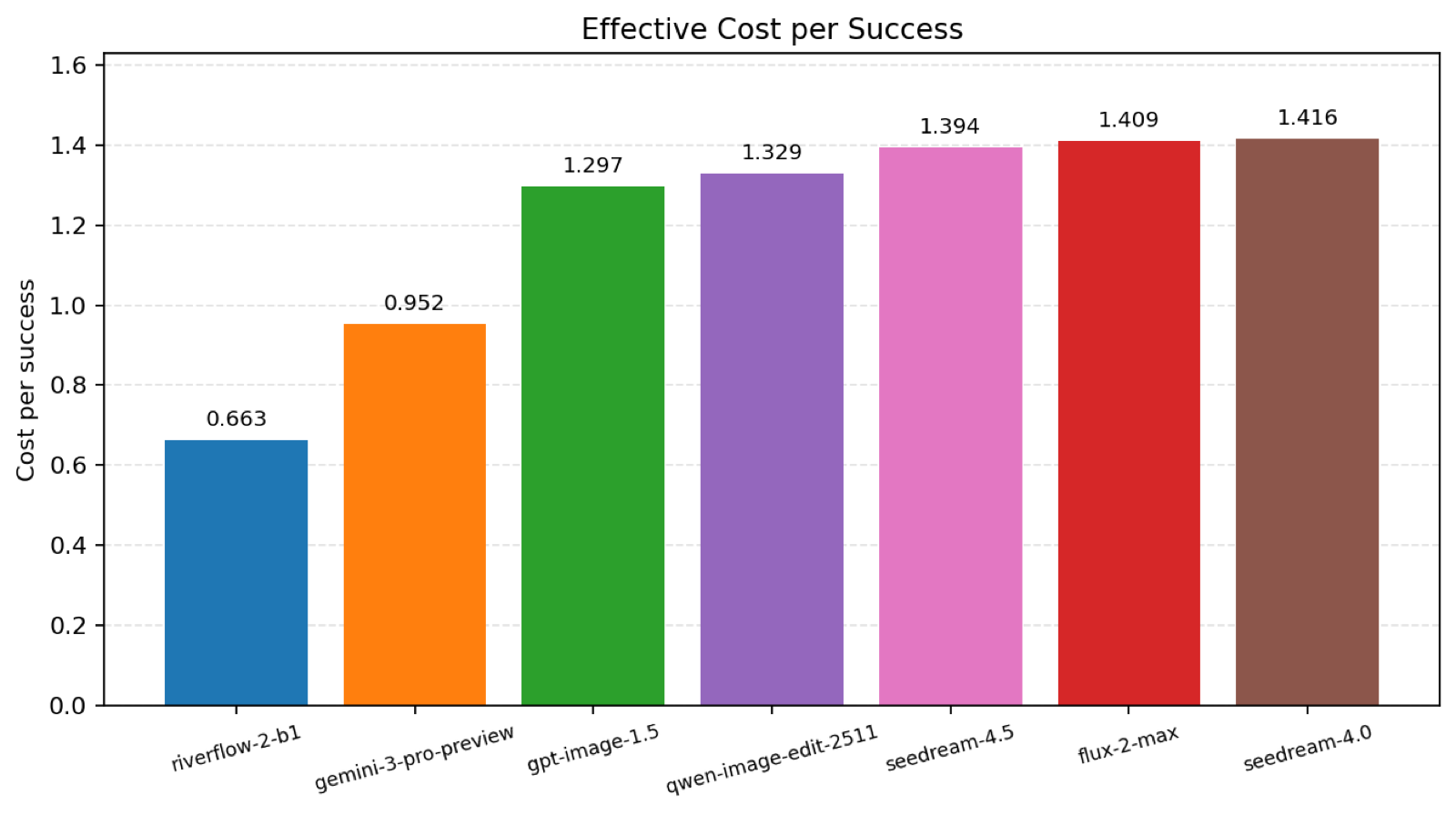}
    \caption{Effective cost (combined).}
  \end{subfigure}

  \begin{subfigure}{0.98\textwidth}
    \centering
    \includegraphics[width=\linewidth,height=0.22\textheight,keepaspectratio]{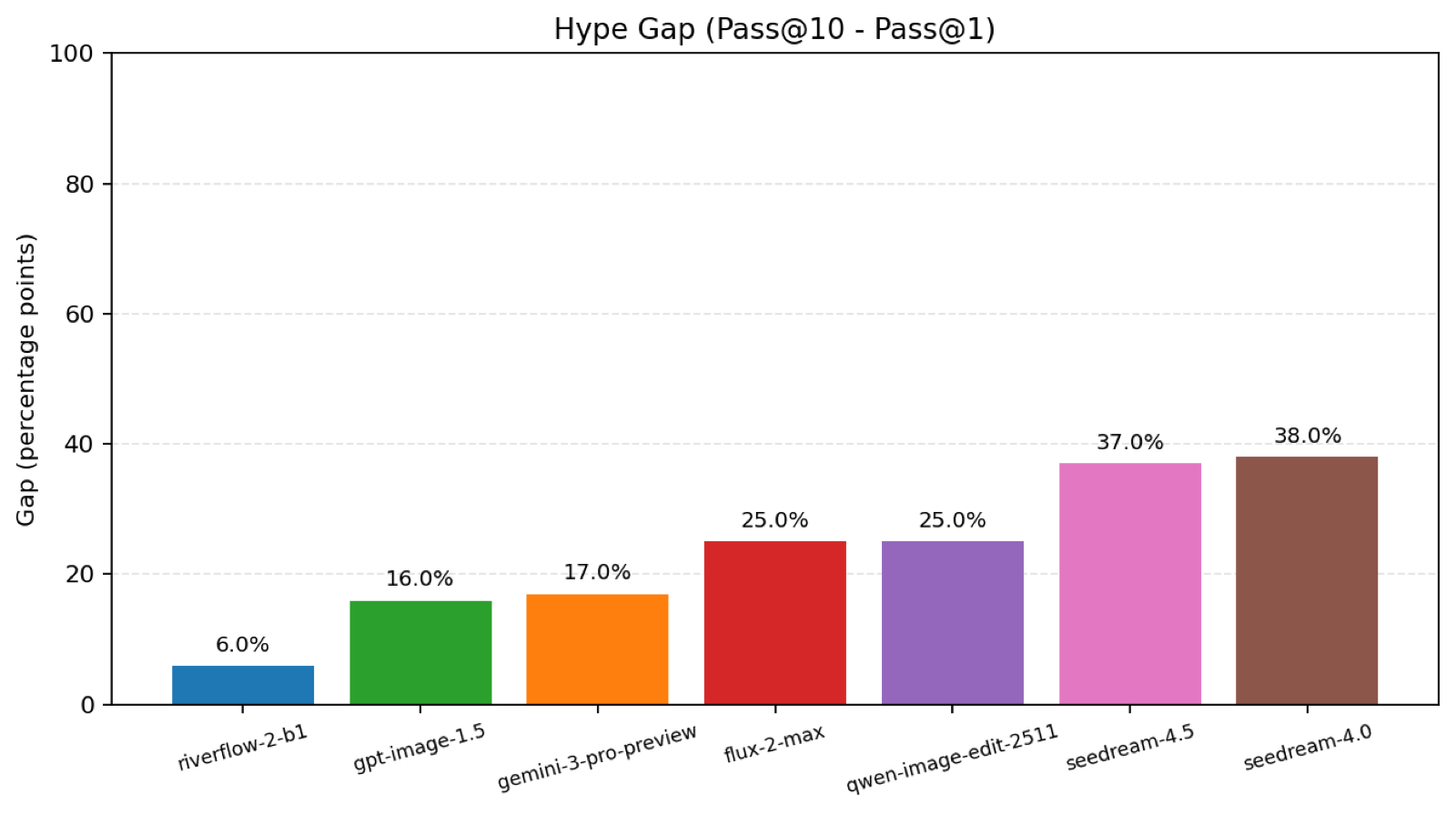}
    \caption{Hype gap (combined).}
  \end{subfigure}
  \caption{Combined split summary charts (human majority labels; VLM judge used only as a check).}
  \label{fig:combined-charts}
\end{figure*}

Table~\ref{tab:combined-results} summarizes the combined split, while
Figure~\ref{fig:combined-charts} visualizes pass rates, pass-at-10 reliability,
expected attempts, effective cost, and hype gap. The gap between pass rate and
pass@10 illustrates how strongly “best-of” sampling can inflate perceived
performance.

\begin{table}[t]
  \caption{Combined split results summary (human majority labels; VLM judge used only as a check).}
  \label{tab:combined-results}
  \centering
  \small
  \setlength{\tabcolsep}{4pt}
  \begin{tabular}{l c c c c}
    \toprule
    Model & Pass Rate (\%) & Pass@4 (\%) & Expected Attempts & Effective Cost (\$) \\
    \midrule
    riverflow-2-b1 & 82.7 & 90.5 & 1.40 & 0.66 \\
    gemini-3-pro-preview & 63.8 & 79.9 & 1.85 & 0.95 \\
    gpt-image-1.5 & 61.2 & 70.3 & 2.04 & 1.30 \\
    flux-2-max & 45.7 & 63.8 & 2.38 & 1.41 \\
    qwen-image-edit-2511 & 45.4 & 57.4 & 2.48 & 1.33 \\
    seedream-4.0 & 35.6 & 57.4 & 2.64 & 1.42 \\
    seedream-4.5 & 34.4 & 59.9 & 2.63 & 1.39 \\
    \bottomrule
  \end{tabular}
\end{table}

\subsection{Public Dataset Results}
\begin{figure*}[t]
  \centering
  \begin{subfigure}{0.49\textwidth}
    \centering
    \includegraphics[width=\linewidth,height=0.22\textheight,keepaspectratio]{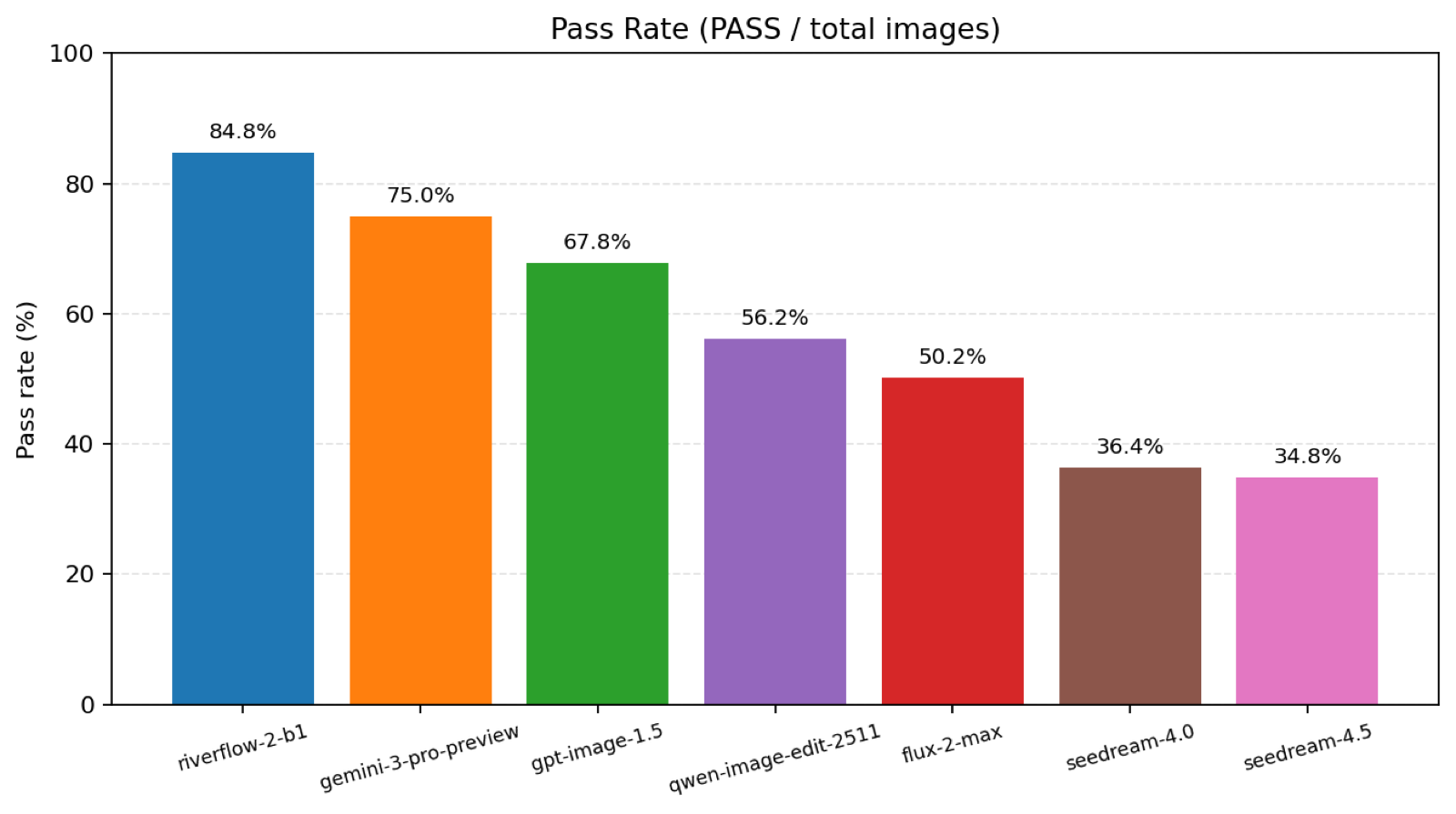}
    \caption{Pass rate (public).}
  \end{subfigure}
  \hfill
  \begin{subfigure}{0.49\textwidth}
    \centering
    \includegraphics[width=\linewidth,height=0.22\textheight,keepaspectratio]{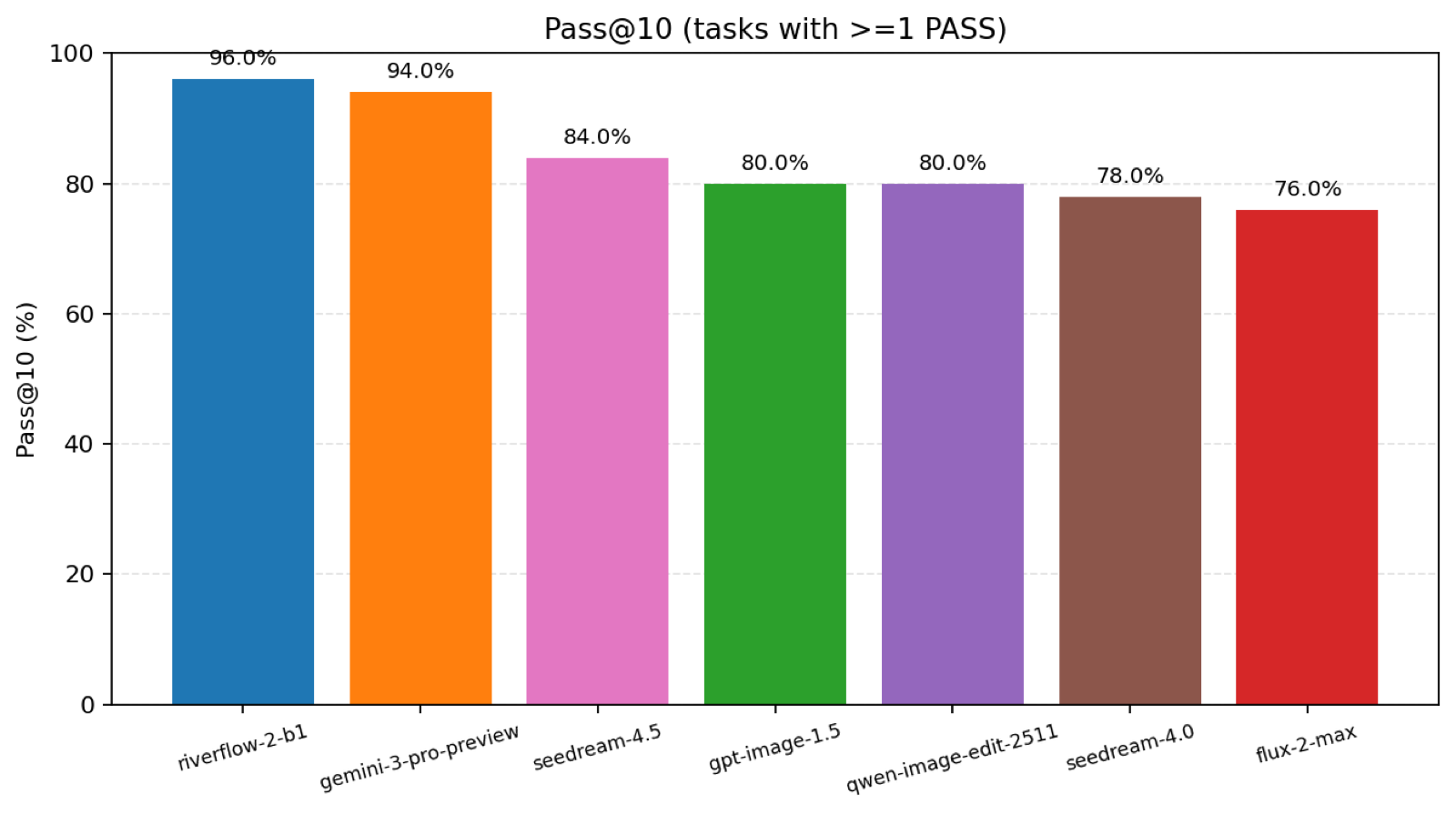}
    \caption{Pass at 10 (public).}
  \end{subfigure}

  \begin{subfigure}{0.49\textwidth}
    \centering
    \includegraphics[width=\linewidth,height=0.22\textheight,keepaspectratio]{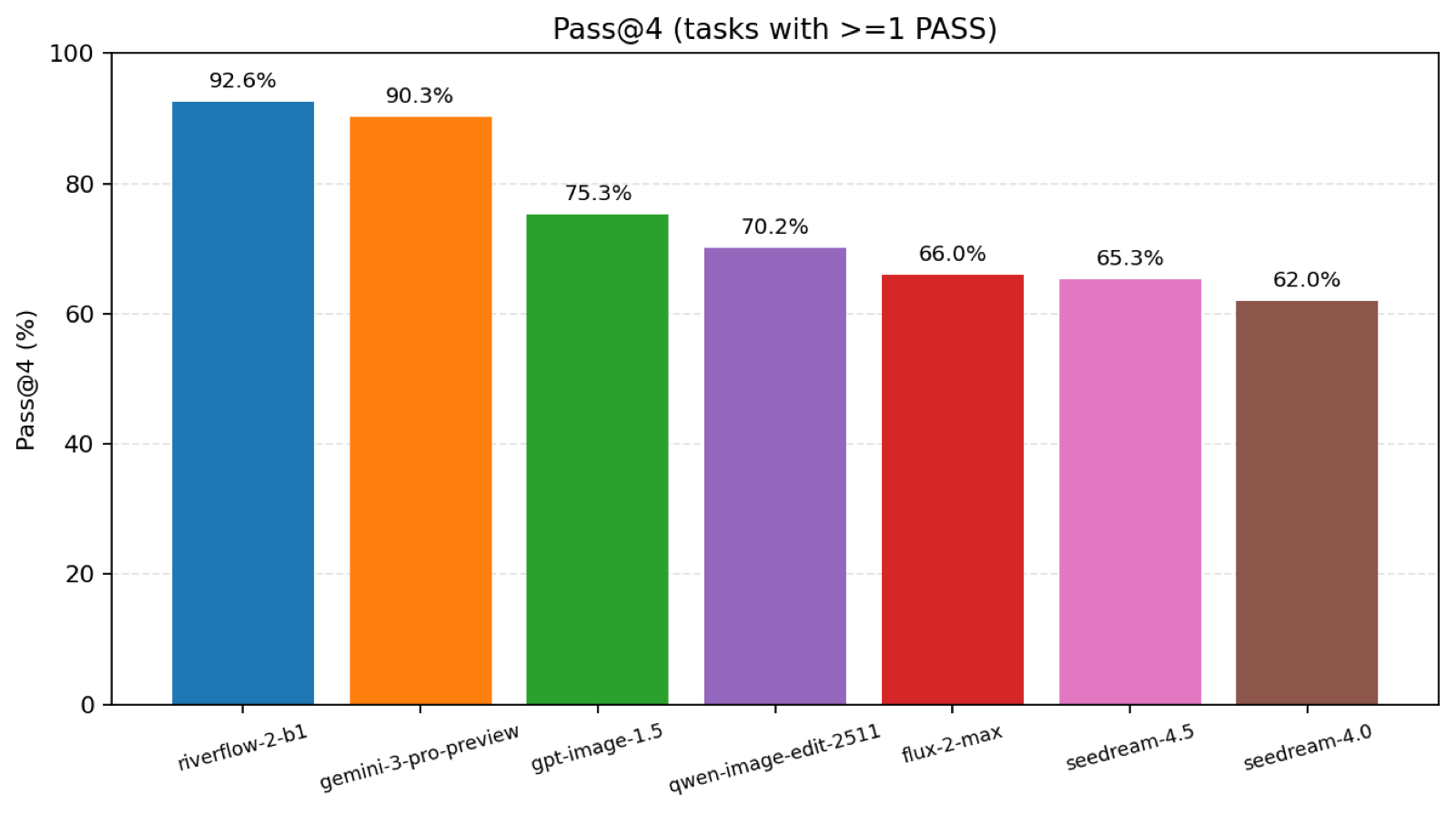}
    \caption{Pass at 4 (public).}
  \end{subfigure}
  \hfill
  \begin{subfigure}{0.49\textwidth}
    \centering
    \includegraphics[width=\linewidth,height=0.22\textheight,keepaspectratio]{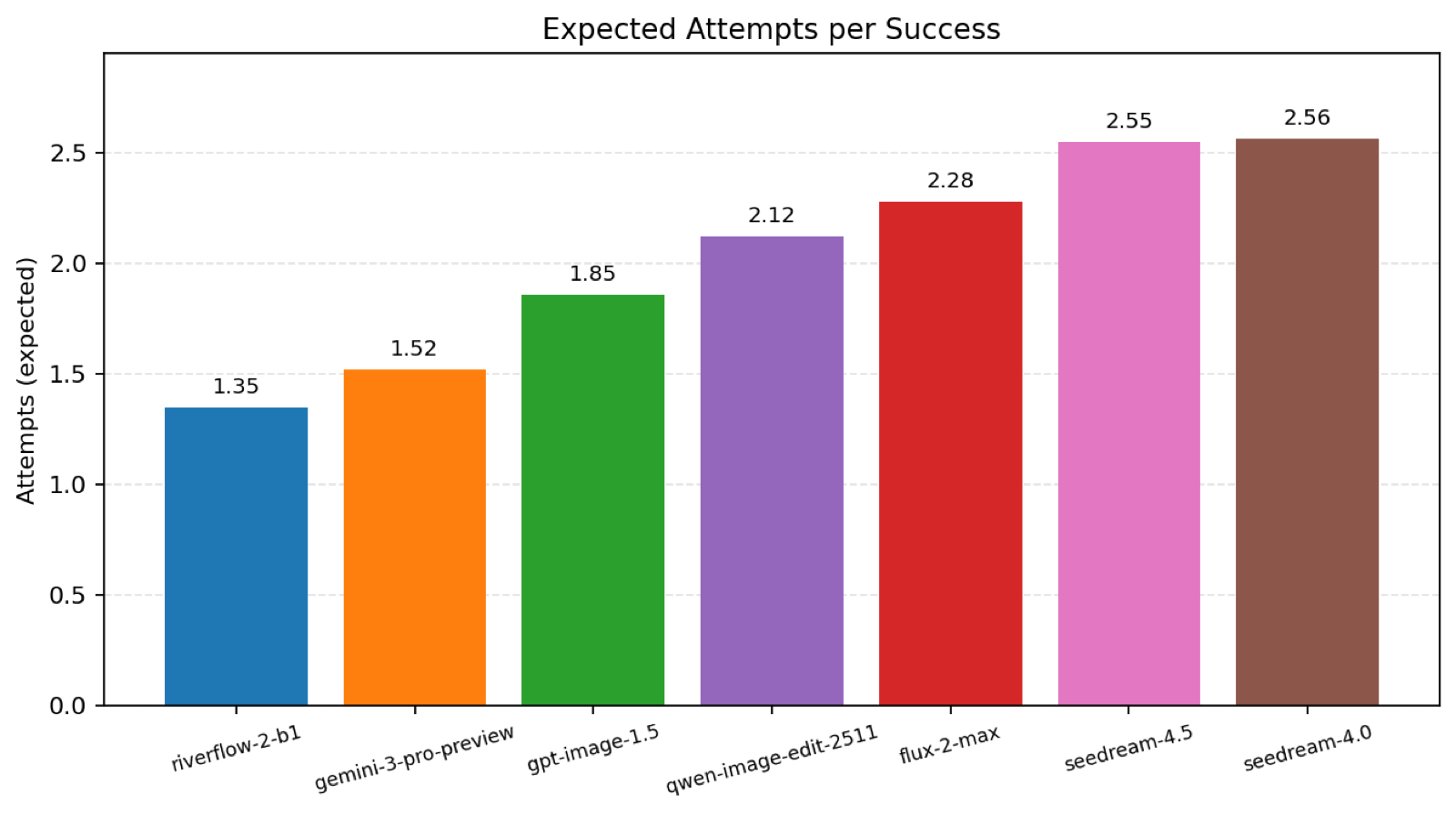}
    \caption{Expected attempts (public).}
  \end{subfigure}

  \begin{subfigure}{0.49\textwidth}
    \centering
    \includegraphics[width=\linewidth,height=0.22\textheight,keepaspectratio]{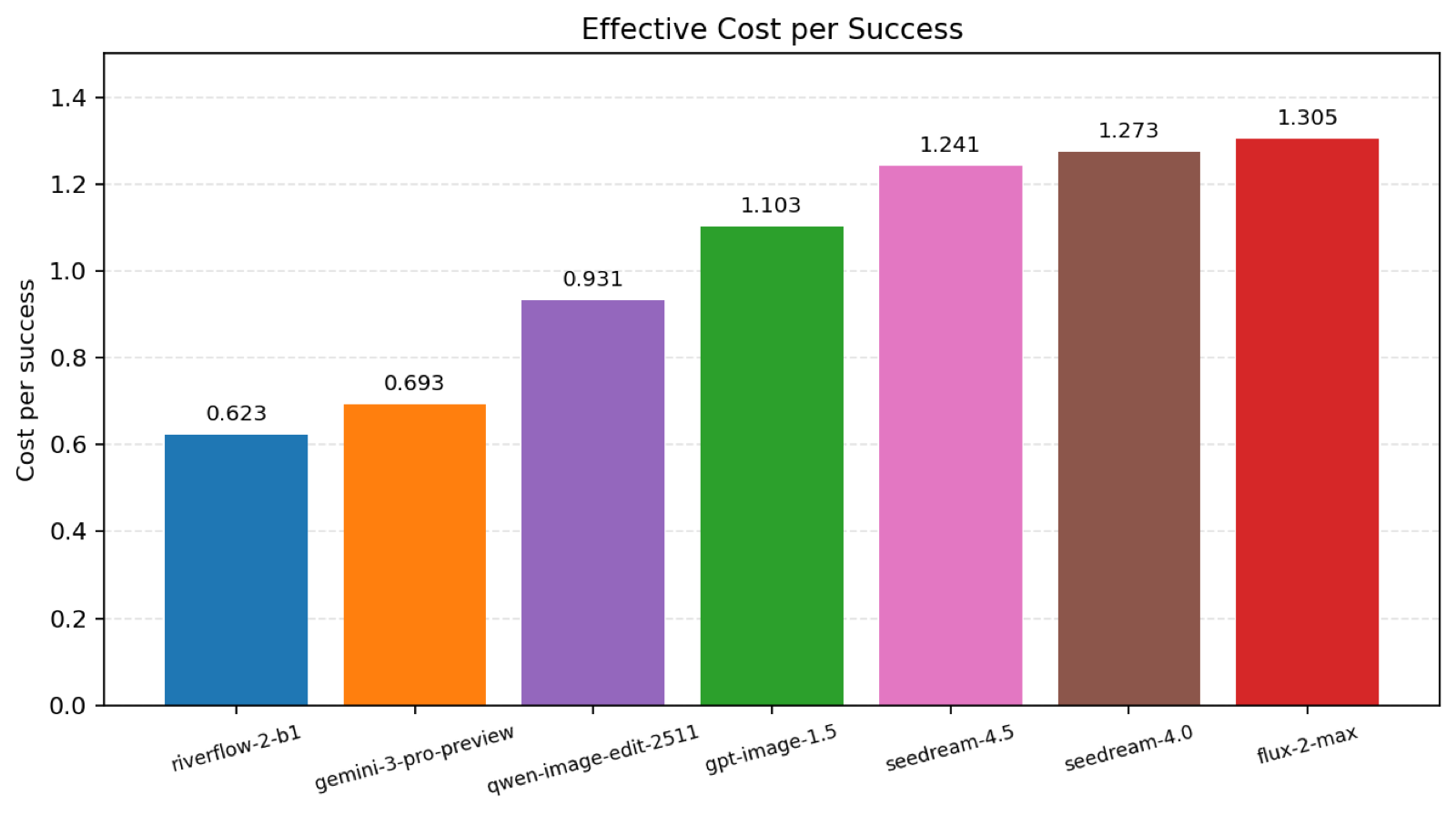}
    \caption{Effective cost (public).}
  \end{subfigure}

  \begin{subfigure}{0.98\textwidth}
    \centering
    \includegraphics[width=\linewidth,height=0.22\textheight,keepaspectratio]{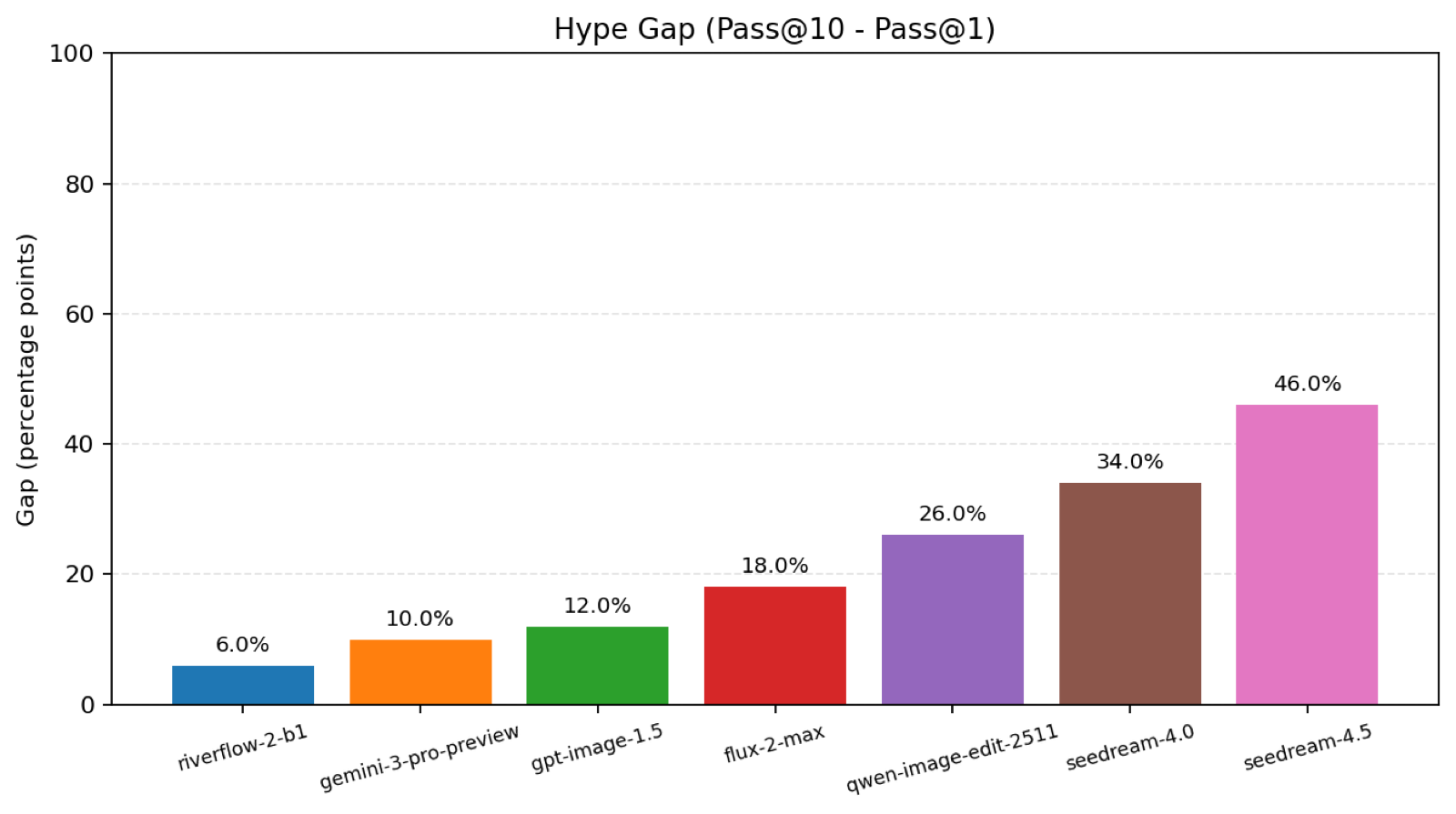}
    \caption{Hype gap (public).}
  \end{subfigure}
  \caption{Public split summary charts (human majority labels).}
  \label{fig:public-charts}
\end{figure*}

Figure~\ref{fig:public-charts} reports the
public split performance across reliability, cost, and hype gap metrics.

\subsection{Private Dataset Results}
\begin{figure*}[t]
  \centering
  \begin{subfigure}{0.49\textwidth}
    \centering
    \includegraphics[width=\linewidth,height=0.22\textheight,keepaspectratio]{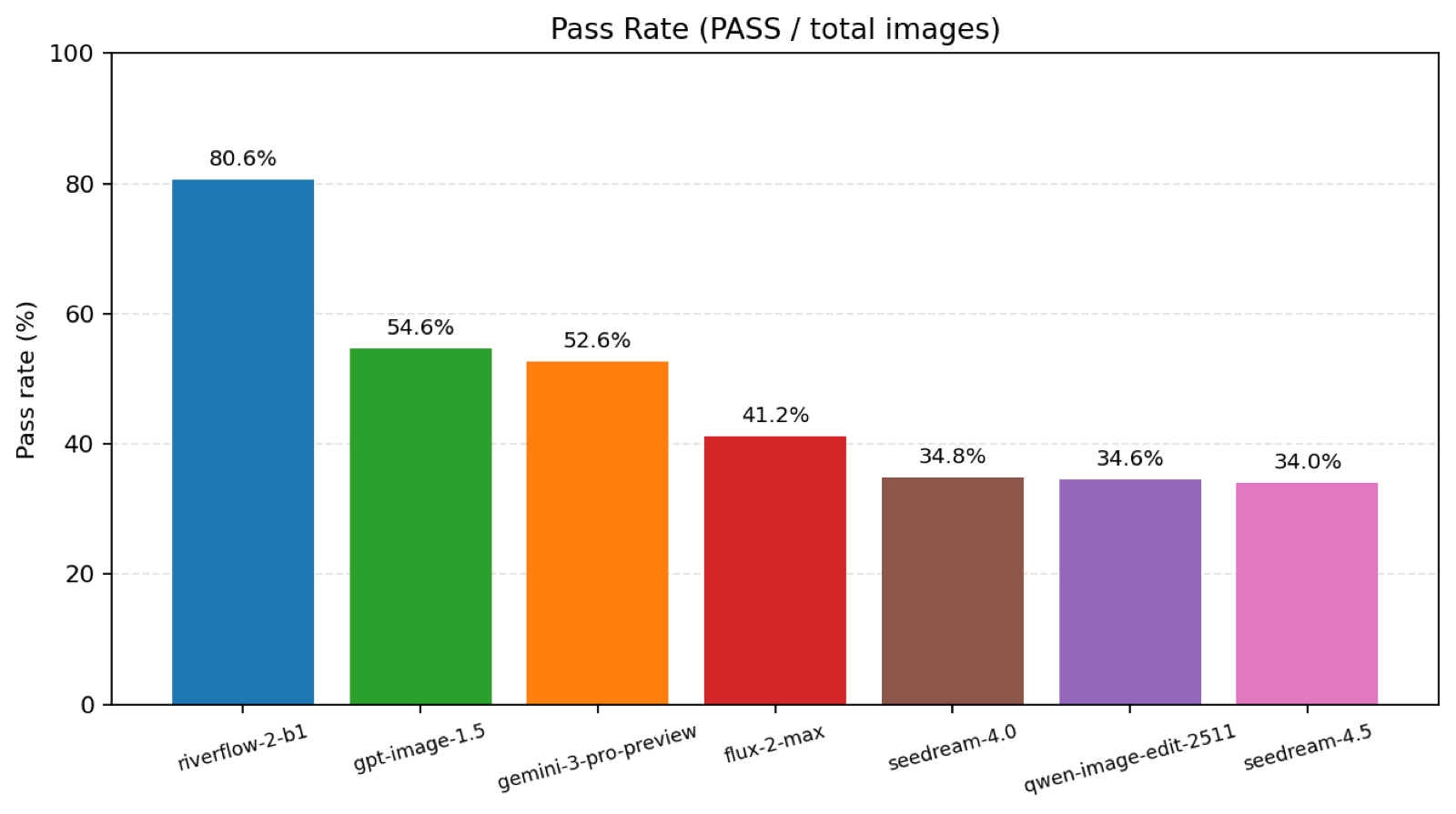}
    \caption{Pass rate (private).}
  \end{subfigure}
  \hfill
  \begin{subfigure}{0.49\textwidth}
    \centering
    \includegraphics[width=\linewidth,height=0.22\textheight,keepaspectratio]{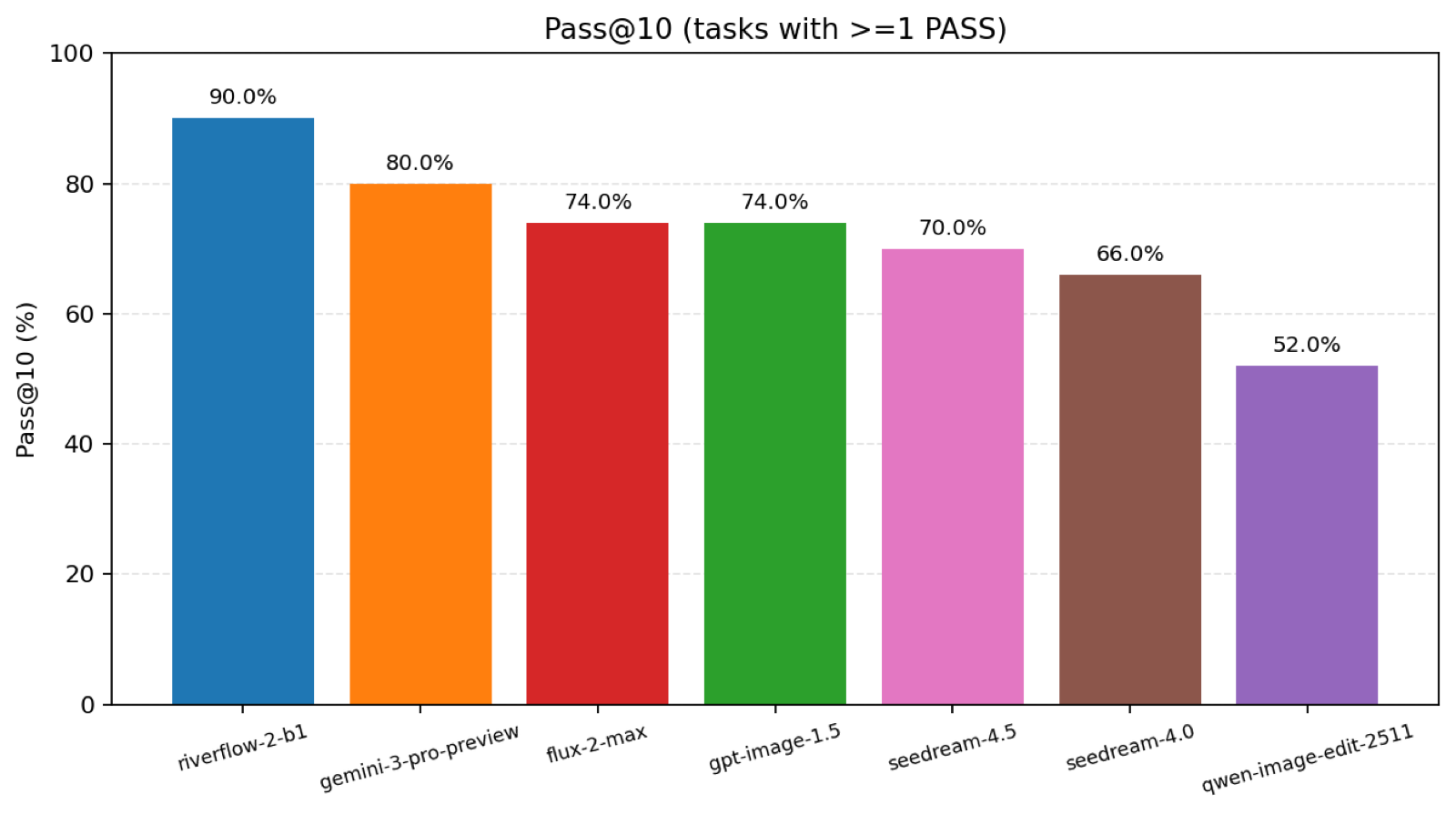}
    \caption{Pass at 10 (private).}
  \end{subfigure}

  \begin{subfigure}{0.49\textwidth}
    \centering
    \includegraphics[width=\linewidth,height=0.22\textheight,keepaspectratio]{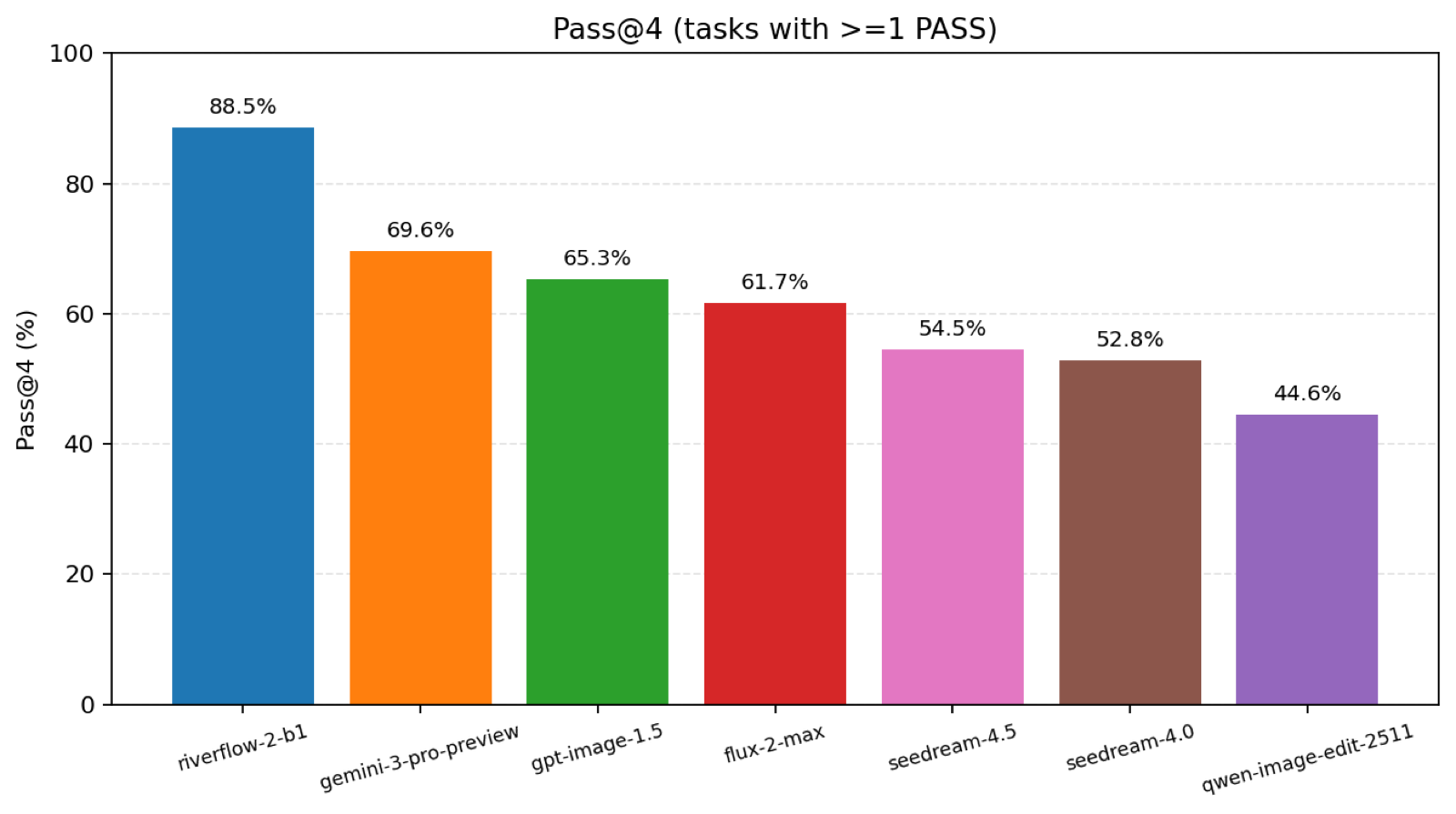}
    \caption{Pass at 4 (private).}
  \end{subfigure}
  \hfill
  \begin{subfigure}{0.49\textwidth}
    \centering
    \includegraphics[width=\linewidth,height=0.22\textheight,keepaspectratio]{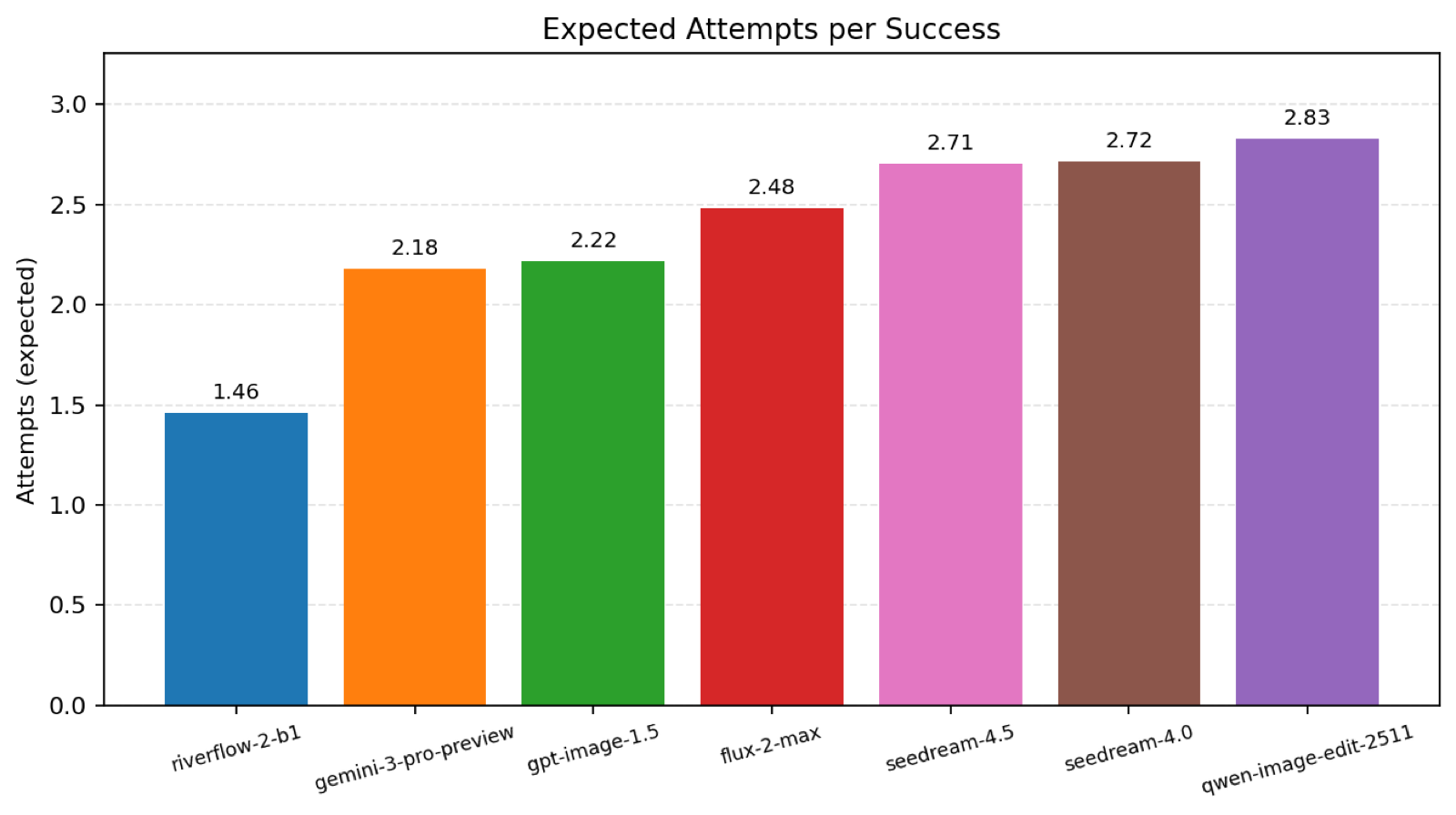}
    \caption{Expected attempts (private).}
  \end{subfigure}

  \begin{subfigure}{0.49\textwidth}
    \centering
    \includegraphics[width=\linewidth,height=0.22\textheight,keepaspectratio]{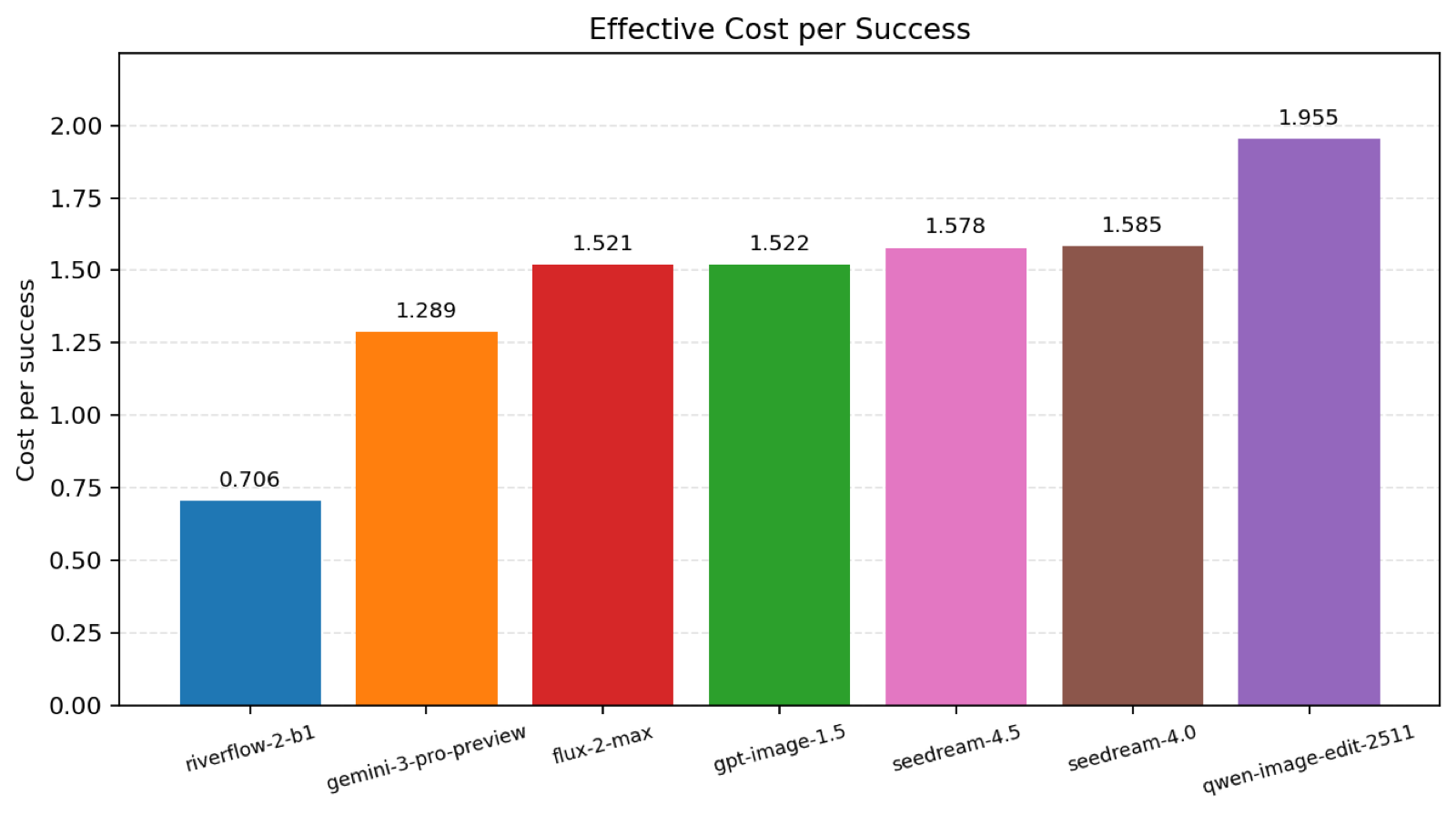}
    \caption{Effective cost (private).}
  \end{subfigure}

  \begin{subfigure}{0.98\textwidth}
    \centering
    \includegraphics[width=\linewidth,height=0.22\textheight,keepaspectratio]{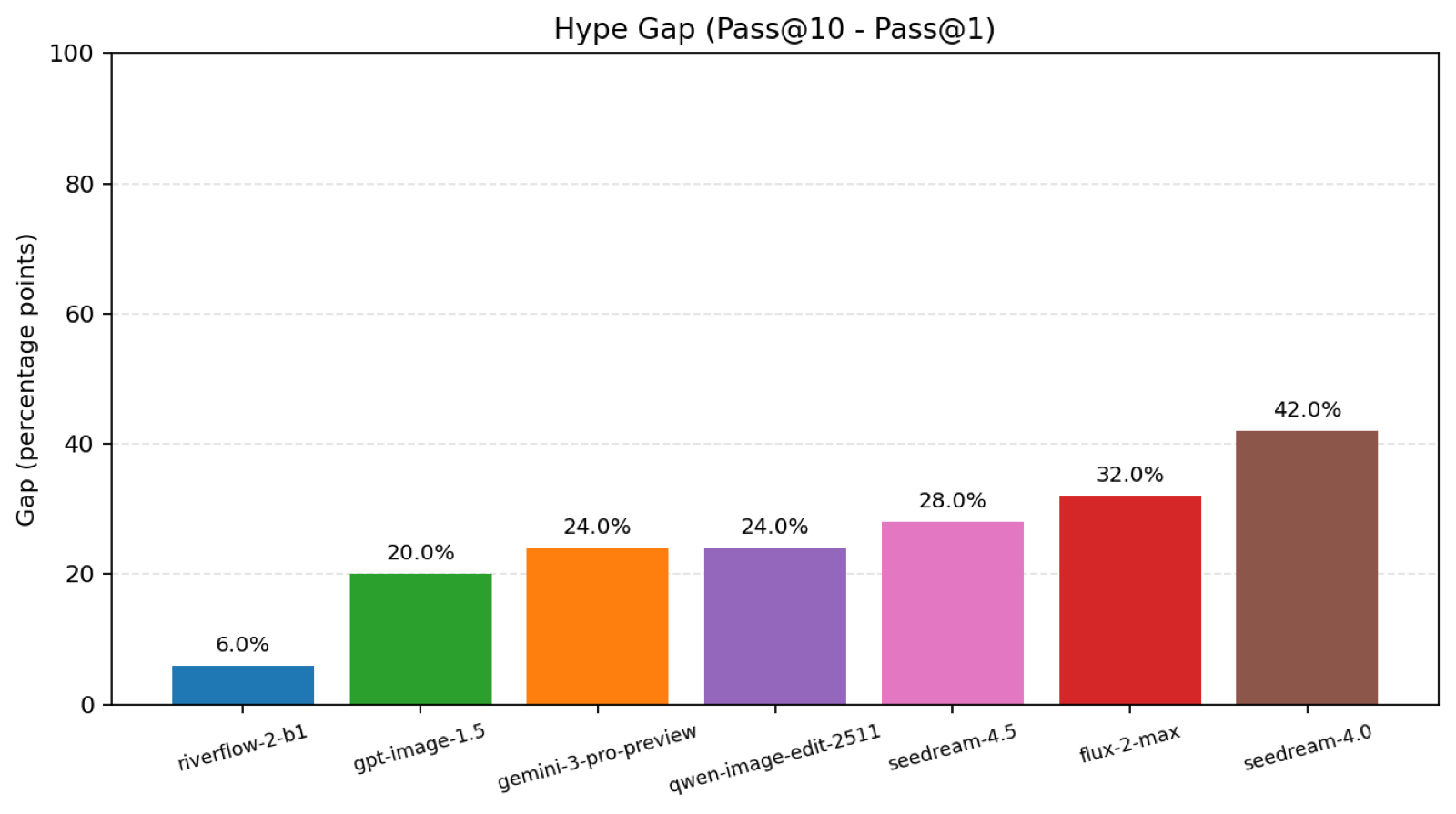}
    \caption{Hype gap (private).}
  \end{subfigure}
  \caption{Private split summary charts (human majority labels).}
  \label{fig:private-charts}
\end{figure*}

Figure~\ref{fig:private-charts} reports the
private split performance across reliability, cost, and hype gap metrics.

\section{Future Work}
HYPE-EDIT-1 is intended as a first reliability-focused benchmark for
reference-based marketing/design edits. We plan a larger follow-on release that
expands both scale and evaluation robustness. Key directions include:
\begin{itemize}
  \item \textbf{Scale and coverage:} expand task count, increase the diversity of
  marketing constraints (layout, typography, product packshots, brand
  compliance), and broaden multi-image edits.
  \item \textbf{Judging robustness:} evaluate with multiple independent VLM
  judges and release a larger human-labeled subset to measure judge correlation
  and bias; additionally, report judge disagreement rates and borderline-case
  analysis.
  \item \textbf{Richer reporting:} provide breakdowns by task type and
  difficulty, and add confidence intervals via bootstrap resampling across
  tasks.
  \item \textbf{Operational sensitivity:} report effective-cost sensitivity to
  review time, labor rate, and retry-cap budgets, and explore alternative user
  workflows such as selecting the best-of-N candidates.
\end{itemize}

\section{Limitations}
HYPE-EDIT-1 focuses on marketing and design edits, so its task distribution does
not cover every possible editing scenario. The repeated-trial protocol assumes
that attempts are independent, which may not perfectly model real-world
conditions. Automated judging checks depend on the reliability of the judge
model and may introduce bias in borderline cases, which is why we rely on human
majority labels and use the VLM as a secondary check.

\section{Broader Impact}
Reliable editing benchmarks can improve transparency in model evaluation and
help users understand the operational cost of deploying image models. However,
broad access to high-quality editing capabilities can be misused for deceptive
content or brand impersonation. Responsible deployment should pair improved
reliability metrics with safeguards and watermarking where appropriate.

\section{Conclusion}
HYPE-EDIT-1 introduces a benchmark that measures both reliability and effective
cost for image editing models using repeated trials on real-world tasks. The
benchmark provides open tooling, a public dataset, and a private evaluation
split to enable consistent, reproducible comparisons. We hope this benchmark
drives model improvements that matter in practical workflows, not just curated
examples.

\FloatBarrier

\appendix
\section{Reproducibility Checklist}
\begin{itemize}
  \item Code and task files are available at \url{https://www.github.com/sourceful-official/hype-edit-1-benchmark}.
  \item Code is licensed under MIT; tasks and reference imagery are licensed under CC BY 4.0.
  \item Public tasks are provided in \texttt{public.json}.
  \item A Gemini 3 Flash (gemini-3-flash-preview) judge example implementation is included.
  \item A human-judge web UI is included for manual evaluation.
\end{itemize}

\end{document}